\documentclass[nohyperref]{article}

\usepackage{microtype}
\usepackage{booktabs} % for professional tables

\usepackage{hyperref}

\usepackage[accepted]{icml2022}

\usepackage{hyperref}
\usepackage{url}
\usepackage{multirow}
\usepackage{graphicx}
\usepackage{caption}
\usepackage{subcaption}
\usepackage{wrapfig}
\usepackage[shortlabels]{enumitem}
\usepackage{colortbl}
\usepackage{mathtools}
\usepackage{pbox}
\usepackage{array}
\usepackage{diagbox}
\usepackage{float}
\usepackage{amsmath}
\usepackage{amssymb}
\usepackage{graphics}
\usepackage{comment}
\usepackage{mathtools}
\usepackage{fancyvrb}
\usepackage{amsthm}
\usepackage{color}
\usepackage{soul}
\usepackage[capitalize,noabbrev]{cleveref}

\newcommand{\bx}{\mathbf{x}}

\newcommand{\cco}{\cellcolor{orange!72}}
\newcommand{\lbc}[2]{\parbox{1.2cm}{\centering #1 \\ (#2)}}

\theoremstyle{plain}

\theoremstyle{definition}

\theoremstyle{remark}

\usepackage[textsize=tiny]{todonotes}

\icmltitlerunning{FOCUS: Familiar Objects in Common and Uncommon Settings}

\begin{document}

\twocolumn[
\icmltitle{FOCUS: Familiar Objects in Common and Uncommon Settings}



\icmlsetsymbol{equal}{*}

\begin{icmlauthorlist}
\icmlauthor{Priyatham Kattakinda}{umd}
\icmlauthor{Soheil Feizi}{umd}
\end{icmlauthorlist}

\icmlaffiliation{umd}{University of Maryland, College Park, MD, USA}

\icmlcorrespondingauthor{Priyatham Kattakinda}{pkattaki@umd.edu}

\icmlkeywords{distributional robustness, image dataset}

\vskip 0.3in
]



\printAffiliationsAndNotice{}  % leave blank if no need to mention equal contribution

% Optional math commands from https://github.com/goodfeli/dlbook_notation.
% \input{math_commands.tex}

\begin{abstract}
Standard training datasets for deep learning often do not contain objects in uncommon and rare settings (e.g., \textit{``a plane on water'', ``a car in snowy weather''}). This can cause models trained on these datasets to incorrectly predict objects that are typical for the context in the image, rather than identifying the objects that are actually present. In this paper, we introduce FOCUS (\textbf{F}amiliar \textbf{O}bjects in \textbf{C}ommon and \textbf{U}ncommon \textbf{S}ettings), a dataset for stress-testing the generalization power of deep image classifiers. By leveraging the power of modern search engines, we deliberately gather data containing objects in common \emph{and} uncommon settings; in a wide range of locations, weather conditions, and time of day. We present a detailed analysis of the performance of various popular image classifiers on our dataset and demonstrate a clear drop in accuracy when classifying images in uncommon settings. We also show that finetuning a model on our dataset drastically improves its ability to focus on the object of interest leading to better generalization. Lastly, we leverage FOCUS to machine annotate additional visual attributes for the entirety of ImageNet. We believe that our dataset will aid researchers in understanding the inability of deep models to generalize well to uncommon settings and drive future work on improving their distributional robustness.

\end{abstract}

\begin{figure*}[t]
\centering
\begin{subfigure}{0.3\textwidth}
\includegraphics[width=\linewidth, trim={0.96cm 0 1cm 0}, clip]{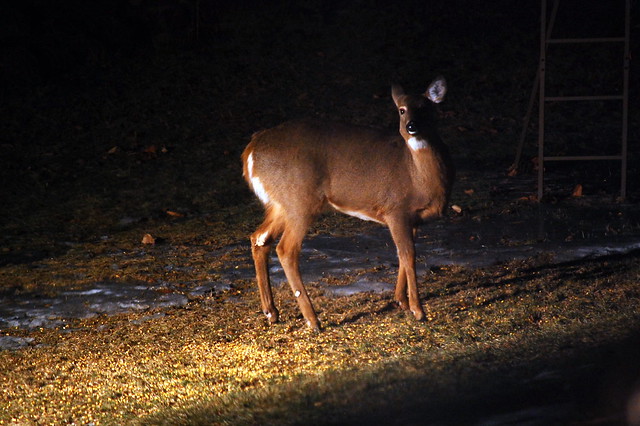}
\caption{\textit{deer at night}}
\end{subfigure}
\begin{subfigure}{0.3\textwidth}
\includegraphics[width=\linewidth, trim={3.5cm 0 7cm 0}, clip]{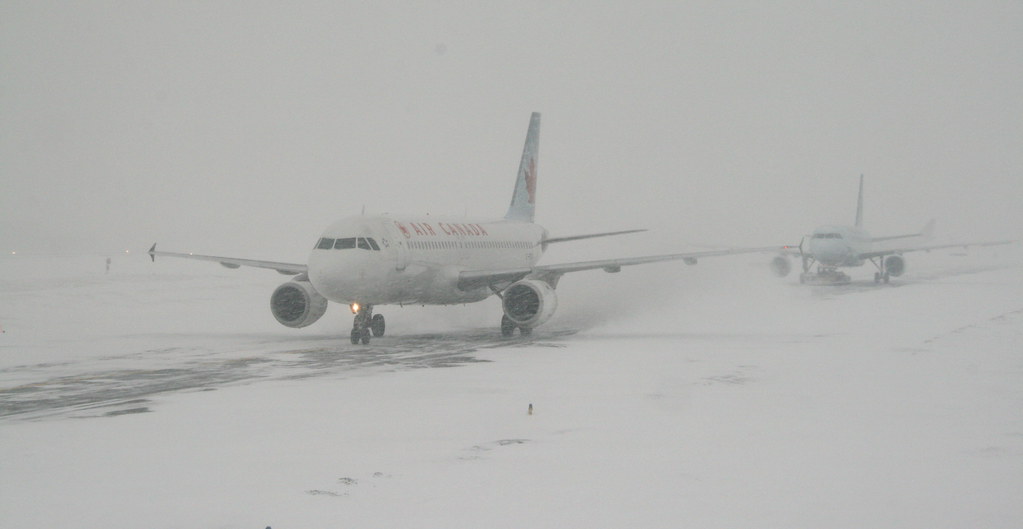}
\caption{\textit{plane in snowy weather}}
\end{subfigure}
\begin{subfigure}{0.3\textwidth}
\includegraphics[width=\linewidth, trim={0 4.4cm 8cm 0}, clip]{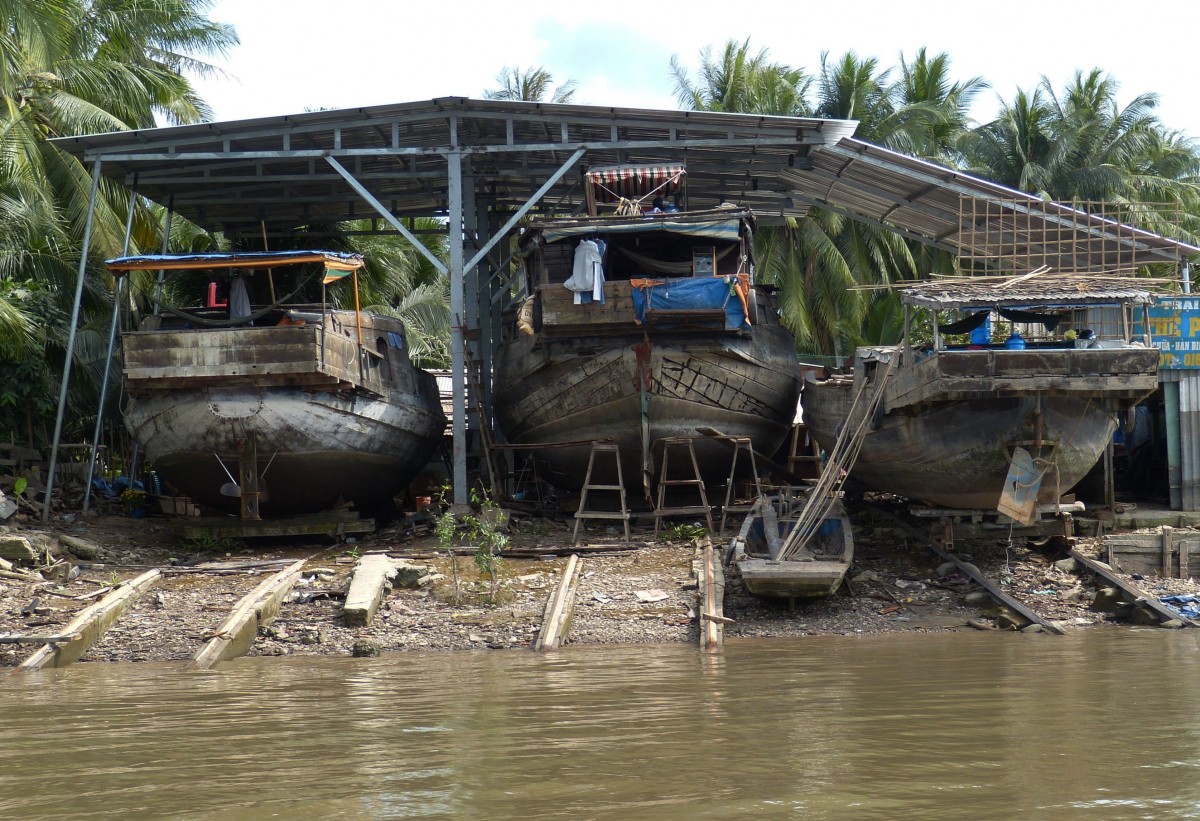}
\caption{\textit{ship indoors}}
\end{subfigure}

\begin{subfigure}{0.3\textwidth}
\includegraphics[width=\linewidth, trim={0cm 0cm 0cm 0}, clip]{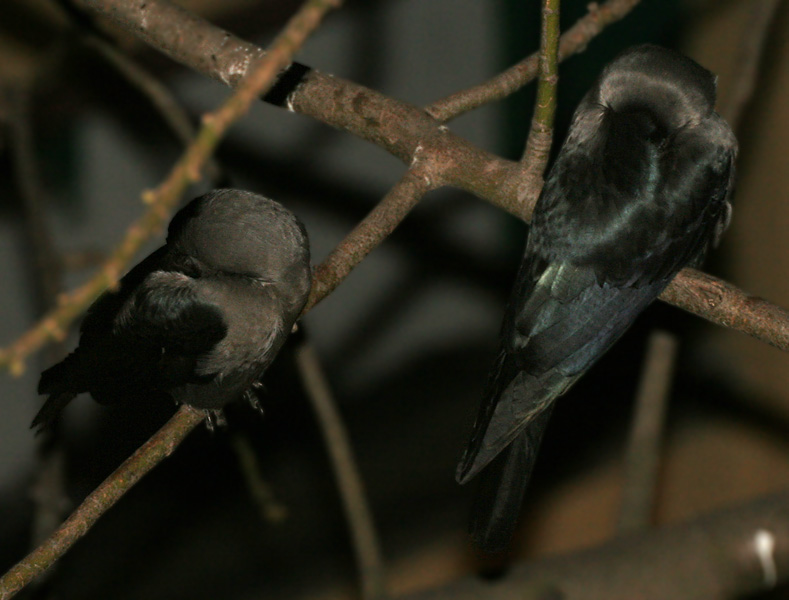}
\caption{\textit{bird at night}}
\end{subfigure}
\begin{subfigure}{0.3\textwidth}
\includegraphics[width=\linewidth, trim={0cm 9cm 0cm 9cm}, clip]{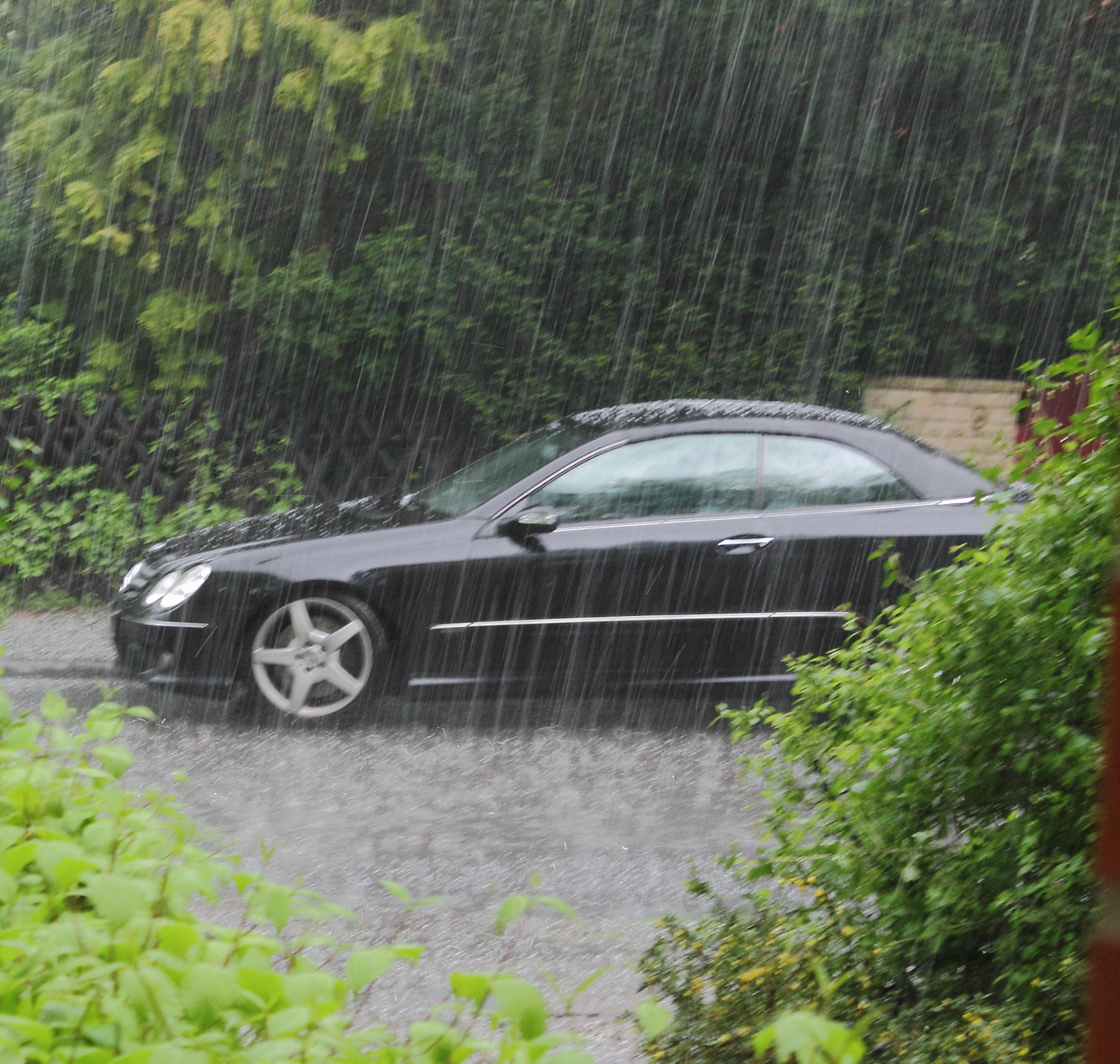}
\caption{\textit{car in rain}}
\end{subfigure}
\begin{subfigure}{0.3\textwidth}
\includegraphics[width=\linewidth, trim={0 0cm 0.2cm 0}, clip]{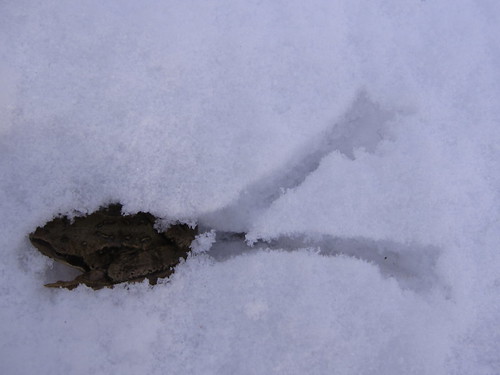}
\caption{\textit{frog in snow}}
\end{subfigure}

\caption{Some uncommon images in the FOCUS dataset. The images in the first column depict uncommon \textit{time of day}, those in the second column depict uncommon \textit{weather}, and the ones in the third column depict uncommon \textit{locations}.}
\label{fig:uncommon-images}

\end{figure*}

\section{Introduction}
Since the remarkable success of AlexNet~\citep{alexnet2012} in the ImageNet Large Scale Visual Recognition Challenge (ILSVRC)~\citep{ilsvrc2015}, deep learning models have been used in a variety of applications ranging from robotics and self-driving cars to stock trading and computational biology. Undoubtedly, large scale datasets such as ImageNet~\citep{deng2009imagenet} deserve much of the credit for the success of deep learning. These datasets typically consist of natural images of objects in some environment. For our purposes, the environment in an image includes all the contextual information surrounding the object in the image. Evidently, objects do not occur independently of their environments. In other words, objects are more likely to be found in some environments than in others (we call these \emph{common settings}). For example, ships are often on water; cars are usually on streets; birds are usually on trees, etc. Search engines are more likely to return images with objects in their common settings when queried for an object alone, i.e., without any additional qualifiers (e.g., just \textit{``deer''} or \textit{``frog''}). As a result, objects in \emph{uncommon settings} are often missing in many of the popular datasets in use today. Therefore, a classifier's performance on these datasets is not indicative of how well it does in novel environments.

To address this issue, we introduce a new dataset containing images both in common and uncommon settings called {\it FOCUS} ({\bf F}amiliar {\bf O}bjects in {\bf C}ommon and {\bf U}ncommon {\bf S}ettings). Our key idea is that modern search engines often return many relevant results even for qualified queries of objects in uncommon settings. For example, searching ``bird indoors" still returns a few relevant images even though this is an uncommon setting. Building on this idea, we collect images of objects in various common and uncommon environments explicitly. FOCUS has around 21K images of ten objects along with annotations for different aspects of the environment in the images including a wide range of locations, weather conditions, and time of day. Depending on the class, we further annotate these environmental settings as \emph{common} or \emph{uncommon}. 

Using FOCUS, we assess the performance of some popular deep learning models with high accuracy on ImageNet, on uncommon settings. We observe that all of these popular models show significant drop in accuracy when tested on objects in uncommon settings. Next, we finetune these models on FOCUS and show clear evidence for substantial improvement in generalization. We also use FOCUS to train classifiers that can detect various visual attributes we consider in FOCUS and use these classifiers to add additional annotations to ImageNet. To the best of our knowledge, FOCUS is the first large-scale dataset of \emph{natural} images with explicit environmental annotations such as locations, weather conditions, and time of day for {\it both} common and uncommon settings. We believe richly annotated datasets such as FOCUS can pave the way to develop models that not only have high accuracy in common settings but are reliable in rare and uncommon settings as well. Our dataset and code for evaluating models on FOCUS are available at \href{https://github.com/priyathamkat/focus}{https://github.com/priyathamkat/focus}.

%We find that generalizing to uncommon time is easier than generalizing to uncommon weather or locations. 

%As we demonstrate in this work, FOCUS can help stress-test deep models and evaluate their generalization power to uncommon settings. 

\section{Related Work}

Deep neural networks are well known to rely on spurious features or ``shortcuts'' for image classification and object detection \cite{geirhos2020shortcut}. \citet{beery2018recognition} demonstrate this phenomenon in the case of detection and classification of animals in uncommon locations; cows are improperly detected or incorrectly classified in beaches. \citet{Sagawa*2020Distributionally} show a similar issue in regard to classification of waterbirds and the color of human hair in CelebA \cite{liu2015faceattributes}. \citet{rosenfeld2018elephant} observe that object detectors may not always detect objects artificially placed into an image. Further, these objects can have a non-local impact, causing the detector to not detect other objects in the image.

In light of the above, test accuracy on datasets like CIFAR~\citep{krizhevsky2014cifar}, ImageNet~\citep{deng2009imagenet}, etc., though crucial, is insufficient to measure the efficacy of deep neural networks. Many datasets have been proposed in the literature for testing the out-of-distribution effectiveness of classification models. \citet{hendrycks2021many} propose three datasets, namely, ImageNet-Renditions, DeepFashion Remixed, StreetView StoreFronts that are designed to test the generalization ability of classifiers to unseen rendition styles, camera view points, and geography, respectively. \citet{barbu2019objectnet} propose ObjectNet, a dataset that provides a far richer variation in the rotation, viewpoint and backgrounds of many object classes in ImageNet. They observe that models trained on ImageNet are 40-45\% less accurate on ObjectNet. Realistic corruptions such as blur, noise, etc., can occur in the real world, for instance, due to camera shake, low light, etc. Thus, training on high-quality clean images may cause classifiers to perform poorly on corrupted images. \citet{hendrycks2019robustness} propose ImageNet-C, a dataset of 15 types of \emph{artifically} corrupted images to systematically study robustness of deep learning models against (synthetic) corruptions. \citet{hendrycks2021many} propose Real Blurry Images, a dataset of 1000 real world blurry images. ImageNet-A \citep{Hendrycks_2021_CVPR} is a dataset of natural, adversarial images which yield drastically low performance on classifiers trained on ImageNet.

In parallel to OOD datasets, many works have been proposed to explicitly find the spurious dependencies on context and/or circumvent them. \citet{singla2021understanding} propose a method for identifying the visual attributes that cause classification failures using the features of an adversarially robust model. \citet{xiao2020noise} propose the Backgrounds Challenge to evaluate how robust models are to (synthetic) changes in backgrounds. \citet{sauer2021counterfactual} propose a generative framework which allows choosing the color, texture and background of a generated image independently. Using this, they generate counterfactual images and show that training on these images improves out-of-distribution robustness. BDD100K\citep{yu2020bdd100k} is a large scale video dataset for driving that covers a wide range of locations, environments and weather conditions. \citet{sparsewong21b} show that training sparse linear models with deep features as inputs results in improved debuggability of neural networks. In a similar vein, 3DB~\citep{leclerc2021three} uses photorealistic simulations to test and debug computer vision models. \citet{beery2018recognition} introduce a dataset of camera trap images. Since, the traps are fixed, the backgrounds in these images are also more or less fixed and hence this dataset provides an ideal testing ground for classification/detection in uncommon contexts.

A large body of influential work proposes models that explicitly use context to improve image classification or object detection performance. \citet{galleguilloscola} use a conditional random field to incorporate spatial and semantic context. \citet{Mottaghi_2014_CVPR} propose a deformable parts based model that uses both local and global context to improve object detection at various scales. \citet{Bell_2016_CVPR} present a novel architecture called the Inside-Outside Net which uses skip pooling to extract context inside the region of interest while the information from outside the ROI is extracted through spatial recurrent neural networks. \citet{choi2011ooc} build an explicit support context model to capture inter-object physical relationships and use it to detect out-of-context objects.  Lastly, \citet{divvala2009context} is an extensive survey on the role of various types of context in object detection.

\setlength{\tabcolsep}{3pt}
\begin{table*}[hbtp]
\caption{A frequency breakdown of the various categories and attributes in the FOCUS dataset. Uncommon settings are highlighted in orange.}
\label{focus-stats-table}
\begin{center}
\begin{tabular}{c | c | c c c c c c c c c c c|}
\hline
& & Truck & Car & Plane & Ship & Cat & Dog & Horse & Deer & Frog & Bird \\
\hline
\multirow{3}{*}{Time of Day}
& Day & 1036 & 2232 & 1498 & 1573 & 1809 & 2415 & 2989 & 1622 & 931 & 2298 \\
& Night & \cco50 & \cco241 & \cco128 & \cco159 & \cco72 & \cco56 & \cco60 & \cco51 & 180 & \cco45 \\
& None & 29 & 212 & 109 & 29 & 941 & 503 & 66 & 34 & 323 & 124 \\
\hline
\multirow{7}{*}{Weather}
& Cloudy & 139 & 301 & 259 & 324 & 50 & 151 & 186 & 95 & 17 & 171 \\
& Foggy & \cco27 & \cco103 & \cco56 & \cco109 & \cco7 & \cco46 & \cco105 & \cco78 & \cco1 & \cco43 \\
& Partly Cloudy & 145 & 310 & 305 & 309 & 111 & 170 & 284 & 175 & 44 & 151 \\
& Raining & \cco10 & \cco92 & \cco19 & \cco7 & \cco11 & \cco7 & \cco14 & \cco2 & \cco3 & \cco17 \\
& Snowing & \cco9 & \cco39 & \cco4 & \cco3 & \cco23 & \cco39 & \cco36 & \cco44 & \cco0 & \cco32 \\
& Sunny & 538 & 956 & 694 & 687 & 639 & 1083 & 957 & 742 & 423 & 1184 \\
& None & 247 & 884 & 398 & 322 & 1981 & 1478 & 533 & 571 & 946 & 869 \\
\hline
\multirow{9}{*}{Locations} 
& Forest & 172 & 378 & \cco178 & \cco79 & \cco123 & \cco247 & 586 & 916 & 142 & 400 \\
& Grass & 326 & 651 & 391 & \cco105 & 517 & 804 & 1142 & 1255 & 274 & 541 \\
& Indoors & \cco31 & \cco259 & \cco125 & \cco5 & 1344 & 790 & \cco93 & \cco10 & \cco39 & \cco40 \\
& Rocks & \cco43 & \cco115 & \cco55 & \cco71 & \cco137 & \cco109 & \cco92 & \cco95 & 239 & 183 \\
& Sand & 295 & 526 & \cco239 & \cco187 & \cco143 & 414 & 662 & \cco214 & 152 & 296 \\
& Street & 708 & 1635 & 389 & \cco103 & 405 & 372 & \cco417 & \cco103 & \cco29 & \cco130 \\
& Snow & \cco88 & \cco255 & \cco110 & \cco140 & \cco127 & \cco298 & \cco206 & \cco290 & \cco7 & \cco176 \\
& Water & \cco46 & \cco211 & \cco296 & 1606 & \cco93 & 403 & \cco296 & \cco122 & 355 & 581 \\
& None & 56 & 96 & 526 & 51 & 372 & 279 & 146 & 49 & 446 & 699 \\
\hline
\end{tabular}
\end{center}
\end{table*}
\setlength{\tabcolsep}{6pt}

\section{FOCUS: A Dataset with Common and Uncommon Settings}\label{focus}

\subsection{Building FOCUS}\label{sec:building-focus}
We use the time of day, weather and the locations depicted in an image to characterize environment in it. Modern search engines are capable of returning relevant results even for uncommon qualified queries of objects (e.g., \textit{``frog indoors''}). We leverage this to collect images of objects in various common and uncommon environments explicitly. Concretely, we query the Microsoft Bing Image Search API with statements of the form \textless\verb_object_\textgreater \, \textless\verb_preposition_\textgreater \, \textless\verb_attribute_\textgreater \, (e.g., \textit{``ship on grass''}). This ensures that we  collect a great variety of uncommon images. We also use synonyms of the object categories and the attributes to increase the number of samples we collect. Note that we only query for images which have a license that permits sharing allowing us to release the FOCUS dataset for the research community.

We collect a total of around 37K images using the above procedure. But the search results are not always accurate; a significant fraction of them do not have the relevant object or if they do, the object is not in the environment mentioned in the search query. In addition, because we query images based on only one attribute, we do not have any information about the other attributes in the images. For instance, we do not know the time of day or the locations in a search result for \textit{``car in rain''}. 

We conduct an Amazon Mechanical Turk study both to improve the accuracy of annotations inferred from the search queries and to collect missing annotations. Images are shown to workers in a series of Human Intelligence Tasks (HITs), and they are asked to annotate the image with the appropriate choice for the different attributes. See the appendix~\ref{app:hit-study} for more details about the design of our HITs.

\subsection{The Data in FOCUS}

The FOCUS dataset is a collection of around 21K images, each annotated with the time of day, the weather condition and the locations in the image. Concretely, our dataset is as follows:

\begin{equation*}
    \{(\bx_i,y_i,t_i,w_i,l_i)\}_{i=1}^{n}
\end{equation*}
where
\begin{align*}
    & \bx_i \text{ is the image} \\
    & y_i \text{ is the object label} \in \{\textit{truck, car, plane, ship, cat, dog,} \\
    & \qquad \qquad \qquad \qquad \quad \, \, \, \, \textit{horse, deer, frog, bird}\}\\
    & t_i \text{ is the time of day} \in \{\textit{day, night, none}\} \\
    & w_i \text{ is the weather} \in \{\textit{cloudy, foggy, partly cloudy, raining,} \\
    & \qquad \qquad \qquad \qquad \, \, \, \textit{snowing, sunny, none}\} \\
    & l_i \text{ are the locations} \subset \{\textit{forest, grass, indoors, rocks, sand,} \\
    & \qquad \qquad \qquad \qquad \quad \textit{street, snow, water, none}\}
\end{align*}

The rationale behind our choices for different attributes is as follows:
\begin{enumerate}
    \item \textbf{Object Label:} We choose to work with the same 10 object classes in CIFAR-10 \citep{krizhevsky2014cifar}. These classes cover 226 (including various birds, animals and vehicles) of the 1000 classes in ImageNet.
    \item \textbf{Time of day:} Most images in standard datasets are captured during the day, when the objects are well lit. In contrast, nighttime images often lack a lot of details and are corrupted by high levels of noise. 
    \item \textbf{Weather:} Our choices of weather are fairly comprehensive and include the raining, snowing and foggy conditions which often produce natural corruptions in images.
    \item \textbf{Locations:} We choose a wide range of locations with a healthy mix between common and uncommon \textit{(object, location)} pairs. Since images are often likely to include a combination of locations, we let the locations attribute of an image to be a \emph{subset} of the above set instead of being exactly one element out of it.
\end{enumerate}

\textit{``none''} is assigned to an attribute if its ambiguous or impossible to determine from the image. In addition, \textit{``none"} is also assigned to $l_i$ of an image if none of the considered locations is in the image. Table~\ref{focus-stats-table} summarizes the number of FOCUS samples for each object in various environments. %Next, we explain how we categorize these attributes to common and uncommon settings for different objects.

\subsection{Common vs. Uncommon Settings}
We consider two sources of uncommon \textit{(object, environment)} pairs:

\begin{enumerate}

\item The pair is uncommon in the real world (e.g., \textit{``ship on grass''}). On the Internet, searching for \textit{``ship''} alone is extremely unlikely to return any images of a ship on grass in the top results. In other words, the rarity of a pair in the real world is reflected in the dataset.

\item The pair is uncommon due to the choice of labels and queries used to construct a particular benchmark. For example, consider the \textit{``plane''} class. ImageNet has two labels corresponding to an airplane: \textit{``warplane, military plane'', ``airliner''}. Neither of these are planes that are usually found on water making \textit{(``plane in water'')} an uncommon, if not a non-existent pair in ImageNet. Seaplanes, however, are not that uncommon in the real world. 
\end{enumerate}

We declare an \textit{(object, attribute)} uncommon if the number of samples corresponding to that pair is low (case 1 above). Additionally, we also declare the \textit{(``plane'', ``water'')} pair as uncommon (case 2 above). Our final choices for uncommon settings are highlighted in orange, in table~\ref{focus-stats-table}. Figure~\ref{fig:uncommon-images} shows some uncommon images from our dataset. 

We believe that assigning ``night'' as uncommon Time of Day and ``snowing'', ``raining'', ``foggy'' as uncommon Weather is reasonable. However, categorizing locations into common and uncommon settings for various objects is more subjective. In order to further justify our assignments, we use Gram matrices to identify out-of-distribution samples~\citep{pmlr-v119-sastry20a}. We first compute the Gram matrix deviations for images in the validation set of ImageNet. Next, we identify the threshold that achieves a 95\% true positive rate (TPR) for these images. Finally, we compute the same deviations for images in FOCUS and note the fraction (table~\ref{tab:ood}) of images with deviations above the threshold. For each class (row), we see that the locations with higher percentages are, generally speaking, tagged as uncommon (highlighted in orange).

%\red{what is the conclusion?}

\setlength{\tabcolsep}{1pt}
\begin{table}[hbtp]
\centering
\caption{Uncommon locations (highlighted in orange) in FOCUS are OOD with respect to ImageNet. Higher percentages (see text for details) indicate that the corresponding images are out-of-distribution.}
\label{tab:ood}
\begin{center}
\begin{tabular}{c | c c c c c c c c c|}
\hline

& Forest & Grass & Indoors & Rocks & Sand & Street & Snow & Water \\
\hline
Truck & 70.9 & 71.3 & \cco70 & \cco78.8 & 70.7 & 66.4 & \cco73.8 & \cco71.4 \\
Car & 69.4 & 68.7 & \cco68.1 & \cco72.6 & 74.5 & 65.2 & \cco68.9 & \cco73.1 \\
Plane & \cco88.2 & 89.4 & \cco91.5 & \cco92.6 & \cco91.1 & 88.6 & \cco85.8 & \cco91.9\\
Ship & \cco62.7 & \cco63.5 & \cco90 & \cco62.6 & \cco62.8 & \cco56.8 & \cco63.9 & 61.8\\
Cat & \cco78.7 & 78.3 & 73.6 & \cco71.8 & \cco78.1 & 71.9 & \cco77.6 & \cco75.9\\
Dog & \cco28.8 & 29 & 27 & \cco30.2 & 32.2 & 26 & \cco29.4 & 33.1\\
Horse & 86.9 & 86.8 & \cco91.7 & \cco87.9 & 88.7 & \cco86.1 & \cco88.6 & \cco89.2 \\
Deer & 81.1 & 81.2 & \cco86.7 & \cco81.5 & \cco78.8 & \cco77.7 & \cco83.6 & \cco77 \\
Frog & 76.2 & 74.4 & \cco75 & 75.6 & 76.1 & \cco76.7 & \cco77.5 & 72.9 \\
Bird & 30.6 & 34.7 & \cco27.5 & 34.4 & 35.5 & \cco36.8 & \cco35.5 & 35.6
\end{tabular}
\end{center}
\end{table}
\setlength{\tabcolsep}{6pt}
We acknowledge that this is by no means the one true way. We facilitate other studies opting to do it in other ways by providing all annotations for the collected samples in our dataset, FOCUS.  

\subsection{Why FOCUS?}
\looseness -1
Consider a dataset for image classification, that is constructed by collecting samples for each object class without controlling for the environment the object is in. If the images are collected from the internet, using a search engine, then the distribution of images in the dataset reflects that of the search results. Barring any bias in the search engine, it is reasonable to expect that the images are an accurate representation of the real world. This means that for a given object, most, if not all, images will depict the object in an environment that is typical or natural for the object to be in (eg., \textit{``ship in water''}). Furthermore, unless the number of samples is exceptionally large, the dataset may not contain some objects in some specific environments (eg., \textit{``plane in water''}).

A deep learning model trained on such a dataset will exploit any correlation between objects and their environments in the images since context is a \textit{shortcut} for identifying the object, especially when the correlation is strong. This can be troublesome when the object of interest is an atypical environment; a model that relies too much on context may fail to identify the object in the absence of context cues it has seen during training. This does not mean that all usage of context is ``spurious'' and undoubtedly, even humans depend on various types of context (namely semantic, spatial, pose, etc.,) \cite{oliva2007role}, especially when the object is occluded or lacks discernible details. However, our argument is that humans' use of context is nuanced; we do not look at the surroundings when we can tell what an object is simply by looking at the object alone. On the contrary, today's deep learning models are indiscriminate in their use of context and therein lies the problem.

Lastly, we believe that FOCUS complements existing datasets such as ObjectNet~\citep{barbu2019objectnet} by controlling for a different set of contexts, namely, illumination, weather, and location. In addition, in order to exercise a rich variation in the control variables, ObjectNet only includes objects that are easily manipulated by humans. However, our procedure for building FOCUS (Section \ref{sec:building-focus}) can be used with any object class.

\section{Evaluating Deep Models on FOCUS}\label{experiments}

It is crucial to ensure that machine learning models are reliable even in rare and uncommon settings especially when they are deployed in safety-critical applications. As an example, consider a self-driving car that infers various attributes about its surroundings using deep learning models. The deployed models may be accurate in $99.99\%$ of cases that occur in common settings (e.g., pedestrian crossing on a sunny day). However, given the vast complexity of the real world, uncommon and corner cases, although rare, are still possible (e.g., a heavily snow covered car cutting in). If a model is unreliable in those uncommon settings, it could make a grave error resulting in loss of life and/or property.

In spirit of the aforementioned, we stress test the generalization power of various deep learning models to uncommon settings using the FOCUS dataset. Specifically, we are considering models that are trained using images close to the mode of the \textit{(object, environment)} distribution (i.e., \textit{common images}) and evaluating them on images that fall more on the tail of the \textit{(object, environment)} distribution (i.e., \textit{uncommon images}).

\begin{figure*}[t]
\centering
  \includegraphics[width=\linewidth]{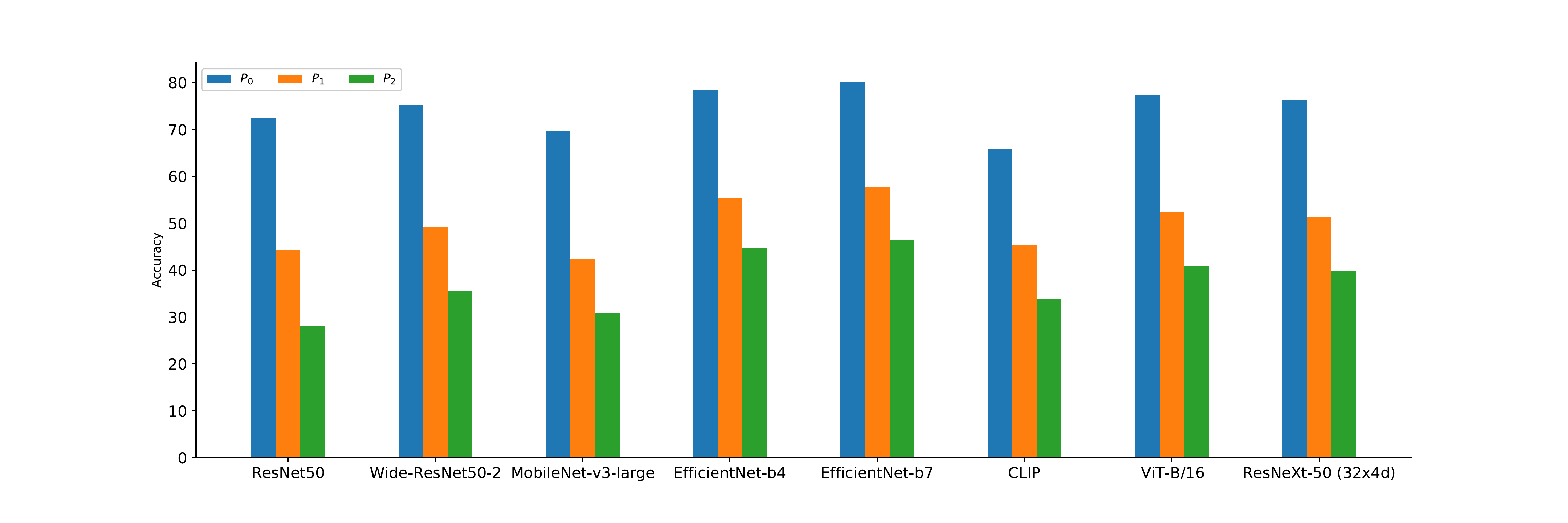}
  \caption{Top-1 classification accuracy for different models on the FOCUS dataset as a function of the partitions $P_i$ }%
  \label{fig:accuracy-vs-num_uncommon}
\end{figure*}
\subsection{Experimental Setup}

\paragraph{Model architectures.} We select some of the most popular deep learning models that have high test accuracy on ImageNet, namely ResNet50~\citep{he2016deep}, Wide-ResNet50-2~\citep{zagoruyko2016wide}, MobileNet-v3-large~\citep{howard2019searching}, EfficientNet-b4~\citep{tan2019efficientnet}, EfficientNet-b7~\citep{tan2019efficientnet}, CLIP~\citep{radford2021learning}, ViT-B/16~\citep{dosovitskiy2020vit}, and ResNeXt-50 (32x4d)~\citep{xie2017aggregated}. For the first three models, we use the pretrained weights provided by PyTorch~\citep{pytorch}. We obtain the weights for both variations of EfficientNet from \url{https://github.com/lukemelas/EfficientNet-PyTorch}. We use the authors' implementation for CLIP (from \url{https://github.com/openai/CLIP}). Following \citet{radford2021learning}, the text inputs to CLIP are of the form ``a photo of \verb^<class>^'', one for each of the 1000 classes in ImageNet. Lastly, we use \verb_timm_~\citep{rw2019timm} for ViT-B/16 and ResNext-50 (32x4d).

\paragraph{Evaluation metrics.}

Since the models we evaluate are pretrained on ImageNet, they output 1,000 probabilities (recall, ImageNet has 1,000 classes). On the other hand, our dataset has only 10 object categories. We resolve this apparent mismatch by first constructing a mapping (denoted by $M$) between the 1000 labels in ImageNet and those in our dataset. Concretely, a label $l_I$ in ImageNet is assigned to a label $l_F$ in our dataset if $l_F$ is a semantic superclass of $l_I$. For example, all the different dog breeds in ImageNet are mapped to the \textit{dog} label in FOCUS. Decidedly, some labels in ImageNet do not have any corresponding labels in our dataset (e.g., \textit{``analog clock'', ``carton'' etc.}). As our dataset has no images from these labels, we declare a misclassification whenever the network predicts a label that is not in the domain of $M$. We say a prediction is correct if the ImageNet label with the highest logit was assigned to the ground truth label in FOCUS. That is, for a sample $(\bx_i,y_i,t_i,w_i,l_i)$ from the FOCUS dataset, let $g(\bx_i)$ be the ImageNet label predicted by a trained network. Then, we have: 
\begin{equation*}
\text{correct prediction} \iff f(\bx_i) \coloneqq M(g(\bx_i)) = y_i.
\end{equation*}

To facilitate the evaluation of the effect of uncommon attributes, we first partition the dataset based on the number of uncommon attributes in images: $P_i$ is a subset of FOCUS samples with $i$ uncommon attributes for $i=0,1,2,3$. Note that $P_0$ denote {\it common} samples while $\bigcup_{i\geq 1} P_i$ denotes {\it uncommon} samples that have at least one uncommon attribute.

\begin{table*}[ht]
\centering
\caption{Sizes of different partitions in FOCUS. $P_i$ is the set of images with $i$ uncommon attributes and $P^A, \, A \subseteq \{t, w, l\} $ is the set of images where the attributes in $A$ are uncommon. Note that $P_0$ constitutes \emph{common} images.}
\label{tab:partition-sizes}

{\renewcommand{\arraystretch}{2}
\begin{tabular}{c c c c c c c c}
\hline
\multicolumn{8}{|c|}{\lbc{Total}{23902}} \\
\hline
\multicolumn{1}{|c}{\lbc{$P_0$}{14144}} & \multicolumn{3}{|c}{\lbc{$P_1$}{5974}} & \multicolumn{3}{|c}{\lbc{$P_2$}{674}} & \multicolumn{1}{|c|}{\lbc{$P_3$}{21}}\\
\hline
 & \multicolumn{1}{|c}{\lbc{$P^{(t)}$}{641}} & \multicolumn{1}{|c}{\lbc{$P^{(w)}$}{448}} & \multicolumn{1}{|c}{\lbc{$P^{(l)}$}{4885}} & \multicolumn{1}{|c}{\lbc{$P^{(t, w)}$}{43}} & \multicolumn{1}{|c}{\lbc{$P^{(w, l)}$}{474}} & \multicolumn{1}{|c}{\lbc{$P^{(t, l)}$}{157}} & \multicolumn{1}{|c|}{\lbc{$P^{(t, w, l)}$}{21}}  \\
 \cline{2-8}
\end{tabular}
}
\end{table*}

We further subdivide $P_i$ into $P^A, \, A \subseteq \{t, w, l\}, \, |A| = i$ where the attributes in $A$ are uncommon (for instance, $P^{(w, l)}$ is the set of all images with two uncommon attributes: weather and location). Table~\ref{tab:partition-sizes} shows the sizes of the different partitions in our dataset.

%\begin{equation}\label{eq:part-acc}
%Acc(P) = \frac{| \{\bx_i \in P \mid f(\bx_i) = y_i\}|}{|P|}. 
%\end{equation}

We then evaluate the classification accuracy of different models in different partitions (referred to as $Acc(P)$).  In an attempt to measure the effect of a single attribute on the accuracy of a model $f$, we define the following generalization gap with respect to an attribute $a$:

\begin{equation}\label{eq:generalization_gap}
\begin{split}
G_a & = \frac{| \{\bx_i \in \text{C}(a) \mid f(\bx_i) = y_i\}|}{|\text{C}(a)|} \\ & - \frac{| \{\bx_i \in \text{UC}(a) \mid f(\bx_i) = y_i\}|}{|\text{UC}(a)|},
\end{split}
\end{equation}

where C($a$) and UC($a$) are the subsets of images in which attribute $a$ is common and uncommon, respectively. Succinctly, $G_a$ is the difference in the classification accuracy between images with a common choice for $a$ and those with an uncommon choice for the same. The larger the $G_a$, the worse the generalization performance of the model on uncommon choices for $a$.

% \begin{figure*}[hbtp]
% \centering
% \includegraphics[width=\linewidth]{images/sec4/resnet_classwise_accuracy_plots.pdf}
% \caption{Classwise top-1 accuracy for ResNet50. The large error bars at $P_2$ are due to insufficient number of samples in this partition. Similar plots for other models are in appendix~\ref{app:model-accs}. }
% \label{fig:resnet-accuracy}
% \end{figure*}

\begin{table}
  \caption{Generalization gap (as in Equation~\ref{eq:generalization_gap}) per attribute for various models. The best gap on each attribute is in boldface.}
  \label{tab:gen-gaps}%
  \begin{tabular}{c | c c c}
Model & $G_t$ & $G_w$ & $G_l$ \\
\hline

ResNet50 & 9\% & 23\% & 20\% \\
Wide-ResNet50-2 & 11\% & 22\% & 19\% \\
MobileNet-v3-large & 12\% & 20\% & 19\% \\ 
EfficientNet-b4 & 7\% & 19\% & 18\% \\
EfficientNet-b7 & 11\% & 18\% & 16\% \\
CLIP & \textbf{0\%} & \textbf{17\%} & \textbf{15\%} \\
ViT-B/16 & 11\% & \textbf{17\%} & 19\% \\
ResNeXt-50 (32x4d) & 8\% & 18\% & 18\%\\

\end{tabular}
\end{table}

% \begin{table*}[t]
% \centering
% \begin{tabular}{c | c  | c c c | c c c}
% Model & $P_0$ & $P^{(t)}$ & $P^{(w)}$ & $P^{(l)}$ & $P^{(t, w)}$ & $P^{(w, l)}$ & $P^{(t, l)}$ \Bstrut \\
% \hline\Tstrut
% ResNet50 & 69.22 & 58.57 & 48.33 & 44.81 & 39.22 & 37.33 & 35.26 \\
% Wide-ResNet50-2 & \textbf{73.04} & 60.42 & 46.82 & 48.19 & 41.18 & 41.86 & 46.15 \\
% MobileNet-v3-large & 67.26 & 54.98 & 46.82 & 42.07 & 27.45 & 38.46 & 35.26 \\
% EfficientNet-b4 & 73.01 & \textbf{64.01}  & \textbf{54.68} & 49.44 & \textbf{47.06} & 45.70 & \textbf{50.00} \\
% EfficientNet-b7 & 72.39 & 59.63 & 49.67 & \textbf{50.89} & 39.22 & \textbf{48.42} & 47.44
% \end{tabular}
% \caption{Top-1 accuracies of different models on various partitions of the dataset. The best accuracy on each partition is in boldface. The models perform the best on the first column corresponding to the common images while the accuracy decreases as the number of uncommon attributes increases.}
% \label{tab:part-acc}
% \end{table*}

\subsection{Results}\label{sec:results}
Figure~\ref{fig:accuracy-vs-num_uncommon} shows the accuracy of different model architectures on different partitions of the FOCUS dataset. We observe that for all the models, the accuracy falls as the number of uncommon attributes increases\footnote{We have not included $P_3$ in this analysis as it has very few samples (only 35).}. Note that both the EfficientNet models have the highest accuracy in all three subsets, with EfficientNet-b7 performing the best among all the models. We postulate that this is because these models have larger input size: 380 for EfficientNet-b4 and 600 for EfficientNet-b7 (see \citet{tan2019efficientnet} for details). However, CLIP has the lowest (best) generalization gap between common and uncommon images (i.e., $A(P_0) - A(\bigcup_{i\geq 1} P_i)$) at  21.6\% and ResNet has the highest gap (i.e., the worst generalization) at 29.8\%. 

Table~\ref{tab:gen-gaps} shows the generalization gap (as in Equation~\ref{eq:generalization_gap}) per attribute for various models. We see that all the gaps are positive, clearly indicating poor generalization ability to uncommon settings. Note that $G_t$ is smaller than both $G_w$ and $G_l$ for all the models. So, these models are not hurt as much by uncommon time (i.e., ``night'') as they are by uncommon weather or location. Additionally, we see that CLIP has the best generalization in uncommon \textit{time of day}, \textit{weather} (tied with ViT-B/16), and \textit{location}.

\begin{figure}[ht]
    \centering
    \includegraphics[width=0.9\linewidth]{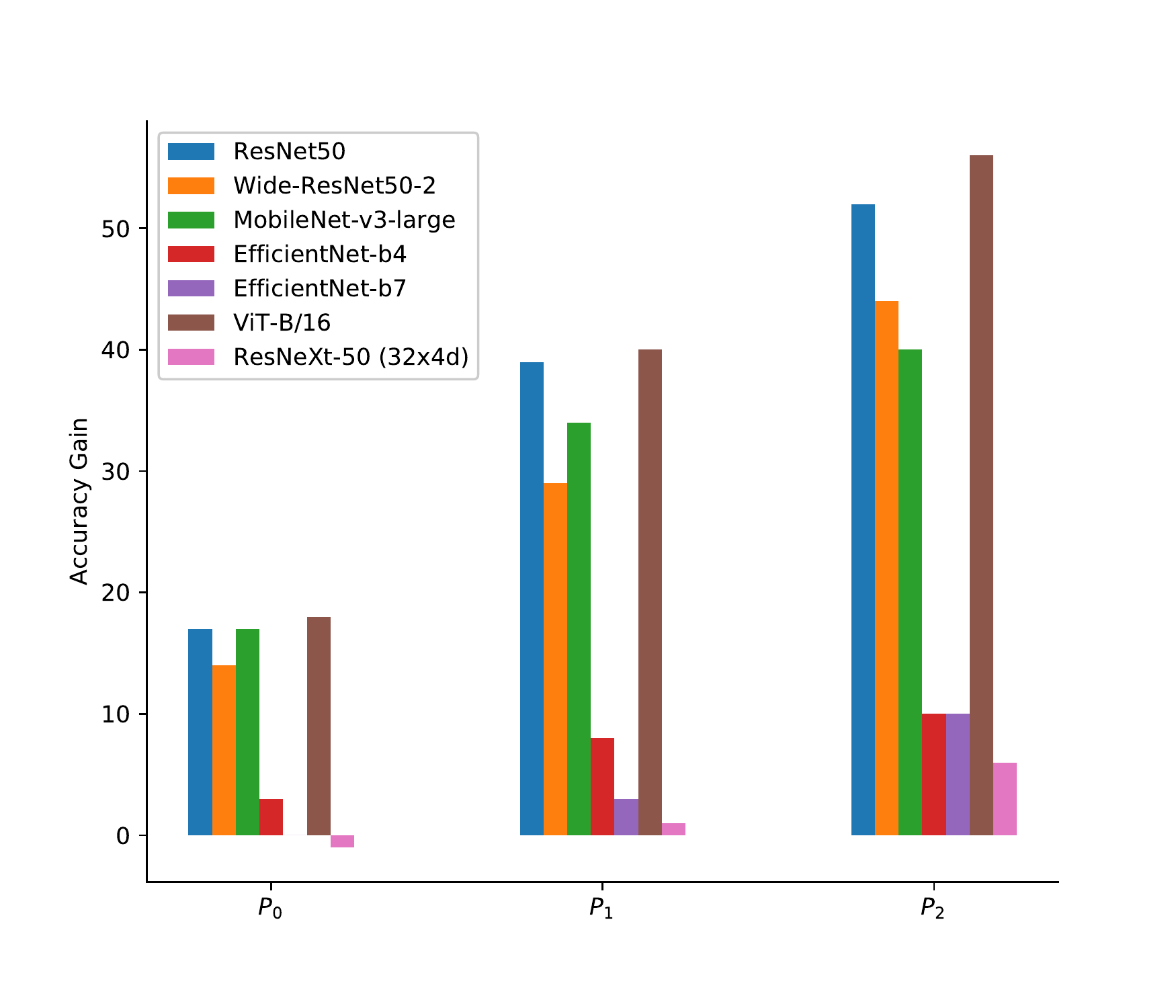}
    \caption{Gains in accuracy of classification of objects on uncommon settings after finetuning on FOCUS.}
    \label{fig:finetuned_accuracies}
\end{figure}

\begin{figure*}[p]
	\centering
	\begin{subtable}[ht]{\textwidth}
        \begin{tabular}{@{\hskip 0.36in}c@{\hskip 0.72in} c@{\hskip 0.9in}c}
        \textbf{Input Image} &  \textbf{Localization before finetuning} & \textbf{Localization after finetuning} \\
    \end{tabular}
    \end{subtable}
	\begin{subfigure}{\textwidth}
	\centering
	\captionsetup{justification=centering}
    \includegraphics[width=0.59\linewidth]{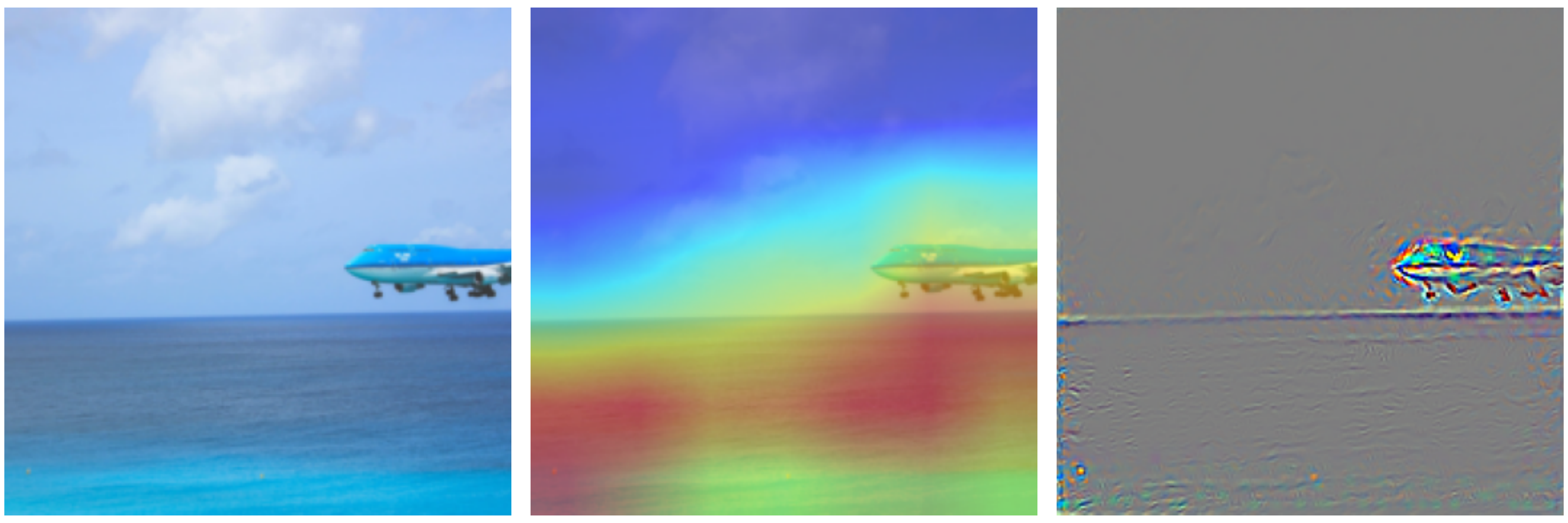}
    \includegraphics[width=0.39\linewidth]{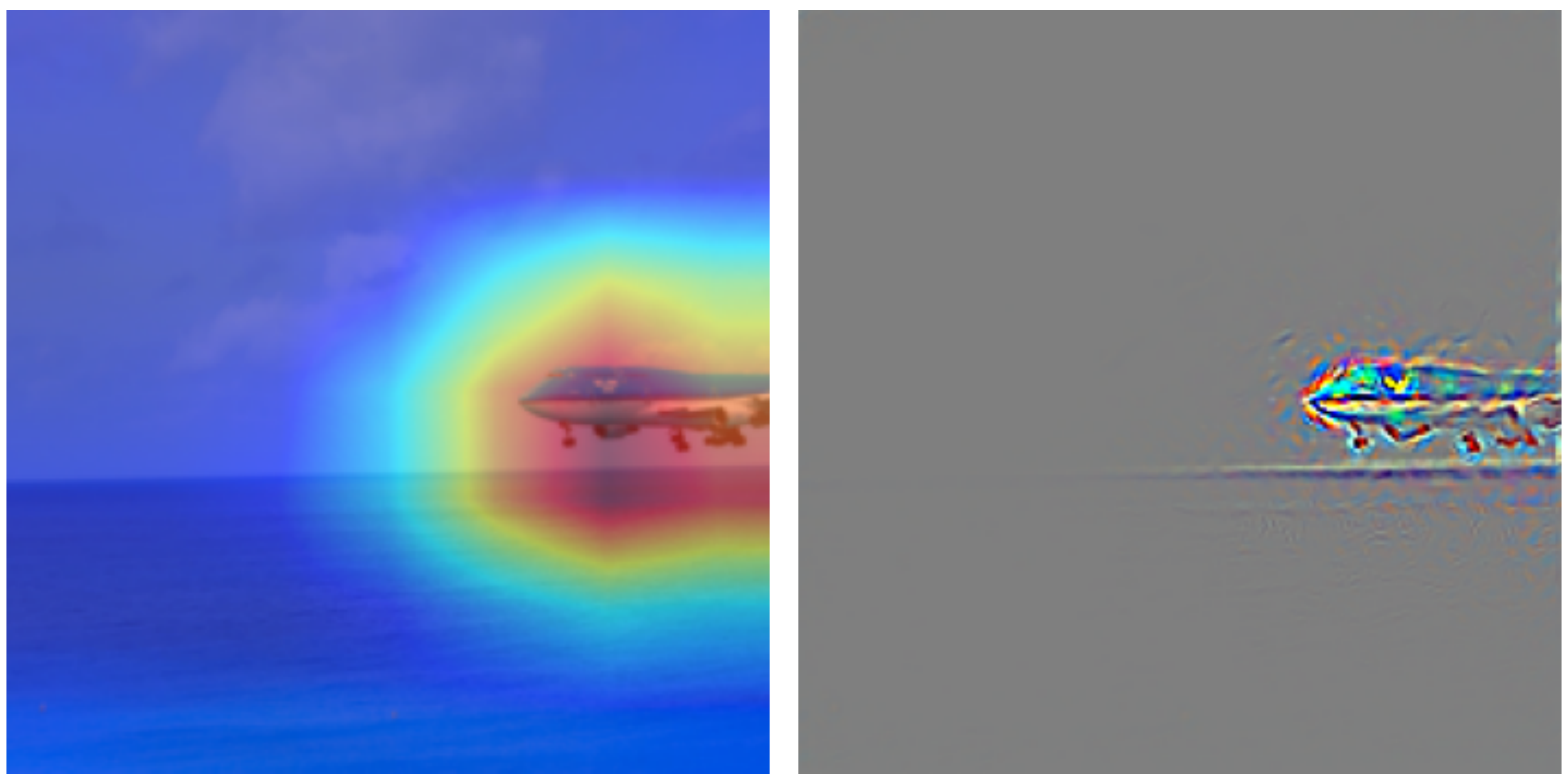}
    \caption{\textit{Plane on Water}. Base model attends to the water and predicts `seashore', while the Finetuned model correctly predicts `warplane'.}
    \label{fig:cam-vis-a}
    \end{subfigure}
    
    \begin{subfigure}{\textwidth}
	\centering
	\captionsetup{justification=centering}
    \includegraphics[width=0.59\linewidth]{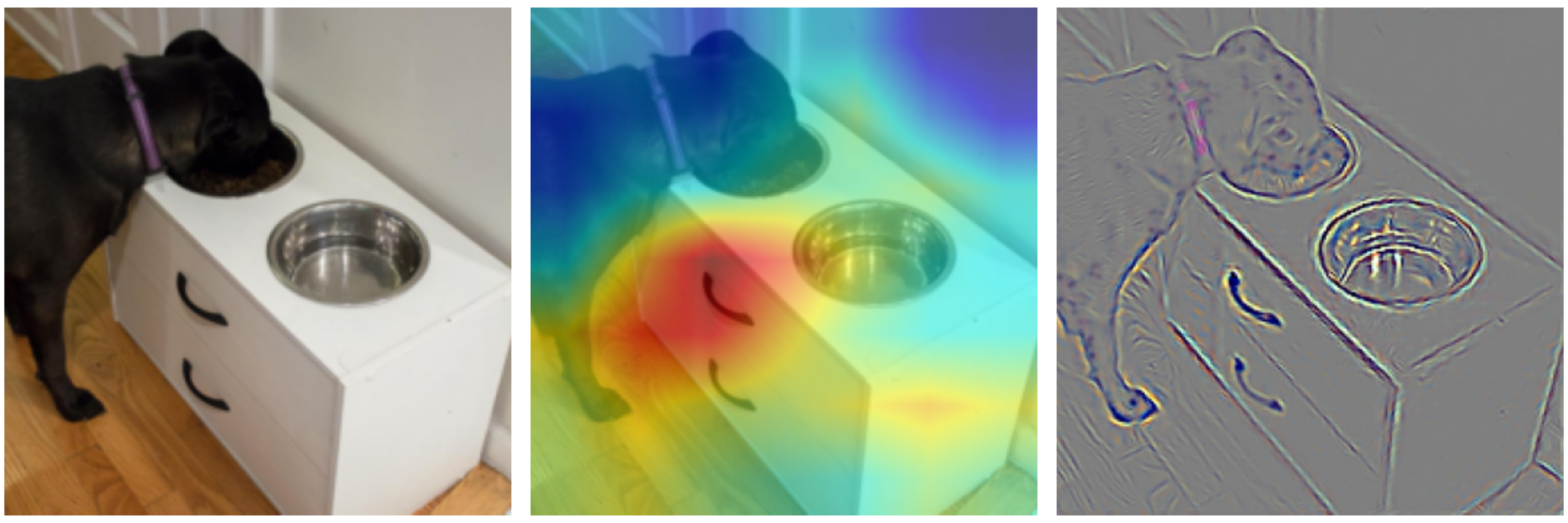}
    \includegraphics[width=0.39\linewidth]{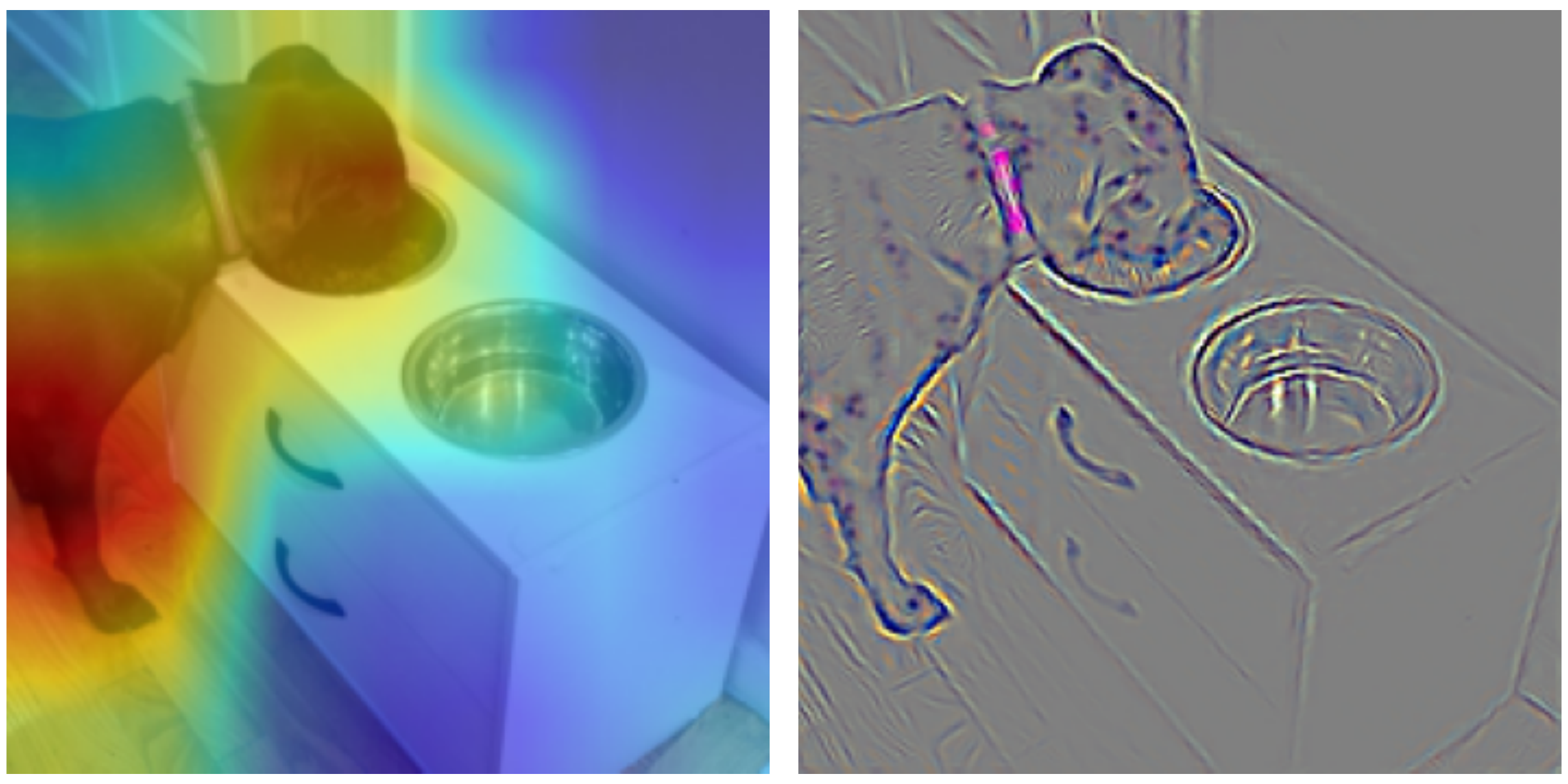}
    \caption{\textit{Dog Indoors}. Base model predicts `dishwasher'. Finetuned model, which localizes on the dog correctly, predicts `Labrador Retriever'. }
    \label{fig:cam-vis-b}
    \end{subfigure}
    
    \begin{subfigure}{\textwidth}
	\centering
	\captionsetup{justification=centering}
    \includegraphics[width=0.59\linewidth]{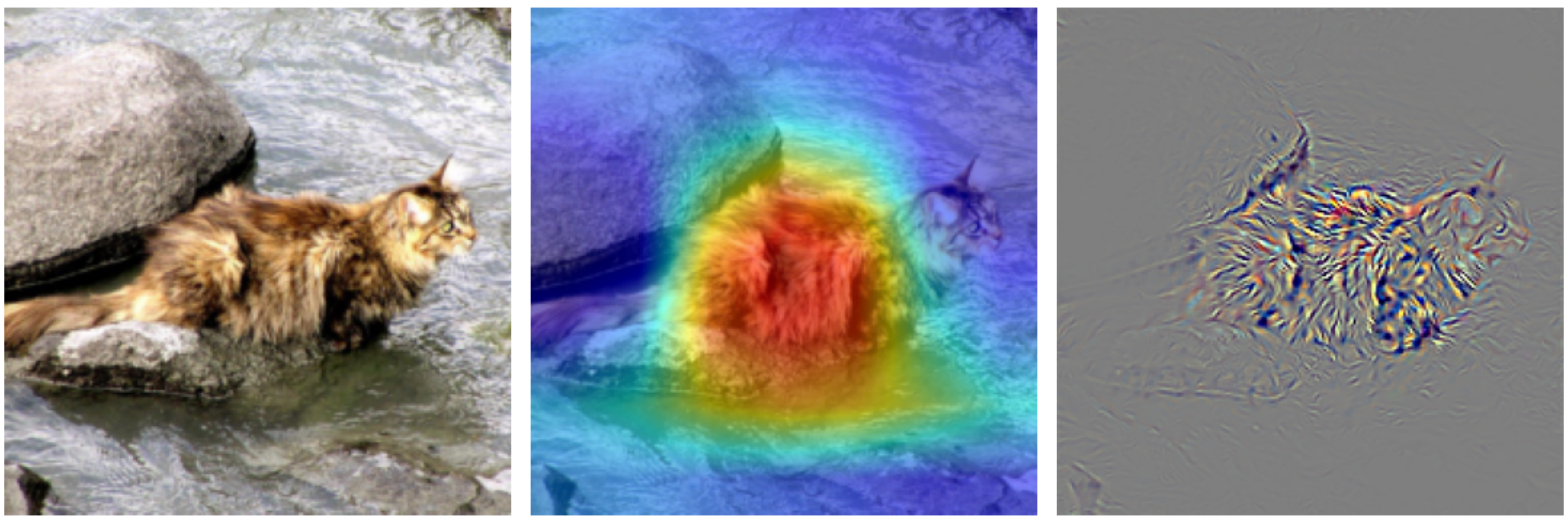}
    \includegraphics[width=0.39\linewidth]{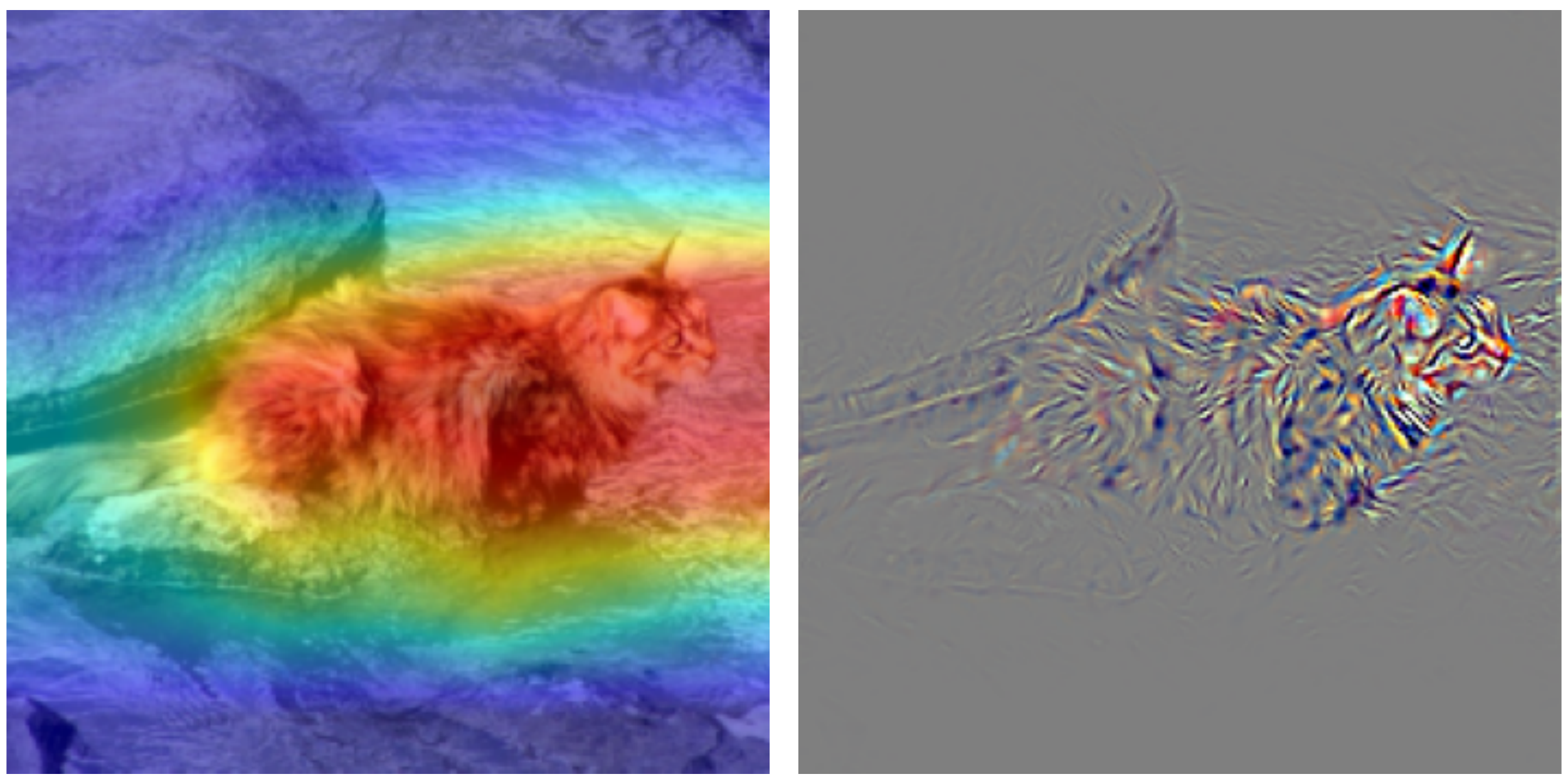}
    \caption{\textit{Cat in Water}. Base model incorrectly predicts `brown bear'. Note the lack of attention on the cat's head. Finetuned model has a more accurate localization map and predicts `tabby cat'.}
    \label{fig:cam-vis-c}
    \end{subfigure}
    
    \caption{Comparison of localization maps of a base ResNet50 model (pretrained on ImageNet) before and after finetuning on FOCUS. In each sub-figure, the first image is the input uncommon image. The second and third images, respectively, show the Grad-CAM~\citep{selvaraju2017grad} and the Guided Grad-CAM of the model \textit{before} finetuning. Finally, the fourth and fifth images show the same for the model \textit{after} finetuning. All Grad-CAM and Guided Grad-CAM images use the predicted class as their target. After finetuning on FOCUS, the model learns to focus more  precisely on the object of interest which leads to substantial increase in classification accuracy (see table~\ref{fig:finetuned_accuracies}).}
    \label{fig:cam-vis}
\end{figure*}

\begin{figure*}[p]
    \centering
    \includegraphics[width=\textwidth]{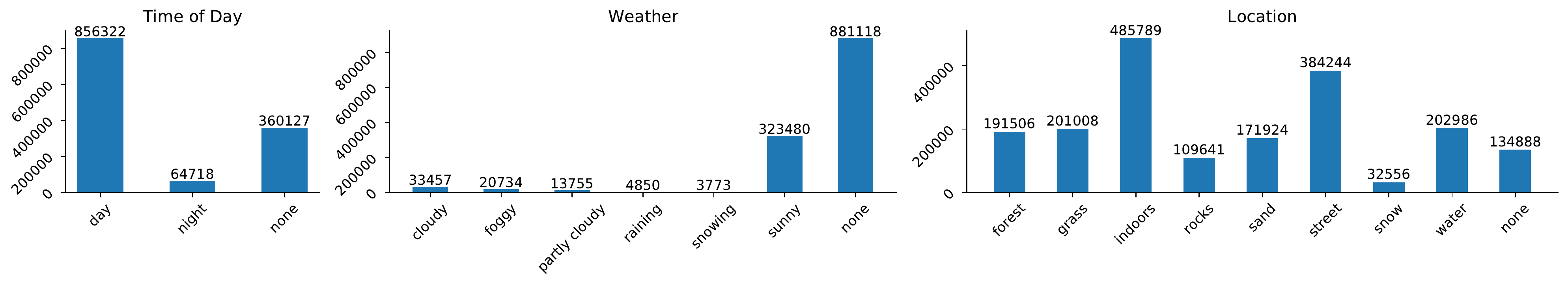}
    \caption{Statistics of various contextual attributes on ImageNet computed by attribute classifiers trained on FOCUS.}
    \label{fig:imagenet-stats}
\end{figure*}

\subsection{Finetuning}
In an attempt to improve classification accuracy in uncommon settings, we finetune the classifiers (except CLIP) we previously tested in subsection \ref{sec:results}, on the FOCUS dataset. We do not fine-tune CLIP on FOCUS because it is a zero-shot classifier and unlike the other models, it does not have a final fully connected layer that can be fine-tuned. We start by randomly splitting the dataset into train and test sets, which are 70\% and 30\% of the dataset in size, respectively. We use SGD with a learning rate of 1e-4 to update the last layer (fully-connected layer) of each model for 10 epochs of the train split. Figure~\ref{fig:finetuned_accuracies} shows the gains in top-1 classification accuracy for each of the model on different partitions of the test set. All models perform better on $P_1$ and $P_2$ after finetuning, with ViT-B/16 having a substantial gain of 40\% and 56\% respectively. This shows that finetuning on FOCUS improves the generalization of classification models and allows them to identify objects in atypical contexts.

Figure~\ref{fig:cam-vis} provides some insight into the effects of finetuning a model on FOCUS. The figure compares the localization maps of a base ResNet50 model that was pretrained on imagenet with those of the finetuned model. Each row illustrates an image from the test split of FOCUS that is misclassified by the base model, but is classified correctly by the finetuned model. It also includes the Grad-CAM~\citep{selvaraju2017grad} localization maps of both the models on that image. In each case, it is evident that the base model is attending heavily to the wrong part of the image. As a result, it incorrectly predicts a class that is closely associated with the background: (a) In figure~\ref{fig:cam-vis-a}, the model is relying on the presence of water to predict `seashore' even though its in the middle of an sea/ocean (b) In figure~\ref{fig:cam-vis-b}, bowl and the handles on the drawer cause the model to predict a dishwasher (c) Lastly, in figure~\ref{fig:cam-vis-c}, the model only looks at the wet, brown fur in water to predict a `brown bear'.  On the other hand, the finetuned model focuses more accurately on the correct object alone, allowing it to identify these objects even though they are present in uncommon settings.

\subsection{Expanding ImageNet with FOCUS}

Although FOCUS has rich sample annotations, its relatively small size prohibits training deep classifiers on it from the scratch. In this section, we show that FOCUS can be effectively used to provide rich annotations for other large-scale and standard image datasets such as ImageNet. To do this, we train three classifiers on FOCUS, one for predicting each of the contextual attributes considered in FOCUS. Specifically, we use a pretrained ResNet50 model as backbone and attach a linear head for $n$-way classification ($n = 3, 7, 9$ for time of day, weather, and location, respectively). To estimate the efficacy of these classifiers on ImageNet, we conducted a separate AMT survey where we ask AMT workers to annotate the contextual attributes for 5,000 randomly chosen images from ImageNet. We find that these classifiers are reasonably accurate: the predictions of the classifier agree with that of the humans  74\%, 71\%, and 75\% of the time, for each of the attributes, respectively. 

Finally, we use the attribute classifiers to provide rich annotations of time, weather and location for all ImageNet samples. Figure~\ref{fig:imagenet-stats} shows the statistics of various attributes in the train split of ImageNet. Interestingly, we observe that some contexts such as inclement weather and images of objects in the dark are severely underrepresented. Given the wide use of Imagenet in the machine learning community, we believe that such richly annotated version of it can facilitate large-scale future works to improve model generalization. To that end, we will release these annotations on ImageNet, along with the FOCUS dataset upon the acceptance of this paper.

%We believe that such rich annotations will facilitate large-scale future works on generalization. To that end, we will release these annotations on ImageNet, along with the FOCUS dataset upon acceptance of this paper.

%In this section, we use FOCUS to add additional annotations for the images in ImageNet. First, 

\section{Conclusion}
In this work, we introduced FOCUS, a dataset that contains images both in common and uncommon settings. FOCUS has around 21K samples annotated by their environmental attributes such as locations, weather, and time of day. Using FOCUS, we evaluated the performance of several popular ImageNet classifiers. These models clearly have reduced performance when classifying images in uncommon settings. We showed that finetuning on our dataset alleviates this issue. We also used FOCUS to machine annotate ImageNet with the attributes mentioned above. We believe that richly annotated datasets such as FOCUS open new directions for the development of deep models that are reliable both in common and uncommon settings.

\section*{Acknowledgements}
The authors would like to thank Yogesh Balaji and Mazda Moayeri for their helpful comments and suggestions. This project was supported in part by NSF CAREER AWARD 1942230, a grant from NIST 60NANB20D134, HR00112090132, ONR grant 13370299 and AWS Machine Learning Research Award.

\bibliography{iclr2022_conference}

\begin{thebibliography}{36}
\providecommand{\natexlab}[1]{#1}
\providecommand{\url}[1]{\texttt{#1}}
\expandafter\ifx\csname urlstyle\endcsname\relax
  \providecommand{\doi}[1]{doi: #1}\else
  \providecommand{\doi}{doi: \begingroup \urlstyle{rm}\Url}\fi

\bibitem[Barbu et~al.(2019)Barbu, Mayo, Alverio, Luo, Wang, Gutfreund,
  Tenenbaum, and Katz]{barbu2019objectnet}
Andrei Barbu, David Mayo, Julian Alverio, William Luo, Christopher Wang, Dan
  Gutfreund, Josh Tenenbaum, and Boris Katz.
\newblock Objectnet: A large-scale bias-controlled dataset for pushing the
  limits of object recognition models.
\newblock \emph{Advances in Neural Information Processing Systems},
  32:\penalty0 9453--9463, 2019.

\bibitem[Beery et~al.(2018)Beery, Van~Horn, and Perona]{beery2018recognition}
Sara Beery, Grant Van~Horn, and Pietro Perona.
\newblock Recognition in terra incognita.
\newblock In \emph{Proceedings of the European conference on computer vision
  (ECCV)}, pp.\  456--473, 2018.

\bibitem[Bell et~al.(2016)Bell, Zitnick, Bala, and Girshick]{Bell_2016_CVPR}
Sean Bell, C.~Lawrence Zitnick, Kavita Bala, and Ross Girshick.
\newblock Inside-outside net: Detecting objects in context with skip pooling
  and recurrent neural networks.
\newblock In \emph{Proceedings of the IEEE Conference on Computer Vision and
  Pattern Recognition (CVPR)}, June 2016.

\bibitem[Choi et~al.(2012)Choi, Torralba, and Willsky]{choi2011ooc}
Myung~Jin Choi, Antonio Torralba, and Alan~S. Willsky.
\newblock Context models and out-of-context objects.
\newblock \emph{Pattern Recogn. Lett.}, 33\penalty0 (7):\penalty0 853–862,
  may 2012.
\newblock ISSN 0167-8655.
\newblock \doi{10.1016/j.patrec.2011.12.004}.
\newblock URL \url{https://doi.org/10.1016/j.patrec.2011.12.004}.

\bibitem[Deng et~al.(2009)Deng, Dong, Socher, Li, Li, and
  Fei-Fei]{deng2009imagenet}
Jia Deng, Wei Dong, Richard Socher, Li-Jia Li, Kai Li, and Li~Fei-Fei.
\newblock Imagenet: A large-scale hierarchical image database.
\newblock In \emph{2009 IEEE conference on computer vision and pattern
  recognition}, pp.\  248--255. Ieee, 2009.

\bibitem[Divvala et~al.(2009)Divvala, Hoiem, Hays, Efros, and
  Hebert]{divvala2009context}
Santosh~K. Divvala, Derek Hoiem, James~H. Hays, Alexei~A. Efros, and Martial
  Hebert.
\newblock An empirical study of context in object detection.
\newblock In \emph{2009 IEEE Conference on Computer Vision and Pattern
  Recognition}, pp.\  1271--1278, 2009.
\newblock \doi{10.1109/CVPR.2009.5206532}.

\bibitem[Dosovitskiy et~al.(2021)Dosovitskiy, Beyer, Kolesnikov, Weissenborn,
  Zhai, Unterthiner, Dehghani, Minderer, Heigold, Gelly, Uszkoreit, and
  Houlsby]{dosovitskiy2020vit}
Alexey Dosovitskiy, Lucas Beyer, Alexander Kolesnikov, Dirk Weissenborn,
  Xiaohua Zhai, Thomas Unterthiner, Mostafa Dehghani, Matthias Minderer, Georg
  Heigold, Sylvain Gelly, Jakob Uszkoreit, and Neil Houlsby.
\newblock An image is worth 16x16 words: Transformers for image recognition at
  scale.
\newblock \emph{ICLR}, 2021.

\bibitem[Galleguillos et~al.(2008)Galleguillos, Rabinovich, and
  Belongie]{galleguilloscola}
Carolina Galleguillos, Andrew Rabinovich, and Serge Belongie.
\newblock Object categorization using co-occurrence, location and appearance.
\newblock In \emph{2008 IEEE Conference on Computer Vision and Pattern
  Recognition}, pp.\  1--8, 2008.
\newblock \doi{10.1109/CVPR.2008.4587799}.

\bibitem[Geirhos et~al.(2020)Geirhos, Jacobsen, Michaelis, Zemel, Brendel,
  Bethge, and Wichmann]{geirhos2020shortcut}
Robert Geirhos, J{\"o}rn-Henrik Jacobsen, Claudio Michaelis, Richard Zemel,
  Wieland Brendel, Matthias Bethge, and Felix~A Wichmann.
\newblock Shortcut learning in deep neural networks.
\newblock \emph{Nature Machine Intelligence}, 2\penalty0 (11):\penalty0
  665--673, 2020.

\bibitem[He et~al.(2016)He, Zhang, Ren, and Sun]{he2016deep}
Kaiming He, Xiangyu Zhang, Shaoqing Ren, and Jian Sun.
\newblock Deep residual learning for image recognition.
\newblock In \emph{Proceedings of the IEEE conference on computer vision and
  pattern recognition}, pp.\  770--778, 2016.

\bibitem[Hendrycks \& Dietterich(2019)Hendrycks and
  Dietterich]{hendrycks2019robustness}
Dan Hendrycks and Thomas Dietterich.
\newblock Benchmarking neural network robustness to common corruptions and
  perturbations.
\newblock \emph{Proceedings of the International Conference on Learning
  Representations}, 2019.

\bibitem[Hendrycks et~al.(2021{\natexlab{a}})Hendrycks, Basart, Mu, Kadavath,
  Wang, Dorundo, Desai, Zhu, Parajuli, Guo, Song, Steinhardt, and
  Gilmer]{hendrycks2021many}
Dan Hendrycks, Steven Basart, Norman Mu, Saurav Kadavath, Frank Wang, Evan
  Dorundo, Rahul Desai, Tyler Zhu, Samyak Parajuli, Mike Guo, Dawn Song, Jacob
  Steinhardt, and Justin Gilmer.
\newblock The many faces of robustness: A critical analysis of
  out-of-distribution generalization.
\newblock \emph{ICCV}, 2021{\natexlab{a}}.

\bibitem[Hendrycks et~al.(2021{\natexlab{b}})Hendrycks, Zhao, Basart,
  Steinhardt, and Song]{Hendrycks_2021_CVPR}
Dan Hendrycks, Kevin Zhao, Steven Basart, Jacob Steinhardt, and Dawn Song.
\newblock Natural adversarial examples.
\newblock In \emph{Proceedings of the IEEE/CVF Conference on Computer Vision
  and Pattern Recognition (CVPR)}, pp.\  15262--15271, June 2021{\natexlab{b}}.

\bibitem[Howard et~al.(2019)Howard, Sandler, Chu, Chen, Chen, Tan, Wang, Zhu,
  Pang, Vasudevan, et~al.]{howard2019searching}
Andrew Howard, Mark Sandler, Grace Chu, Liang-Chieh Chen, Bo~Chen, Mingxing
  Tan, Weijun Wang, Yukun Zhu, Ruoming Pang, Vijay Vasudevan, et~al.
\newblock Searching for mobilenetv3.
\newblock In \emph{Proceedings of the IEEE/CVF International Conference on
  Computer Vision}, pp.\  1314--1324, 2019.

\bibitem[Krizhevsky et~al.(2012)Krizhevsky, Sutskever, and Hinton]{alexnet2012}
Alex Krizhevsky, Ilya Sutskever, and Geoffrey~E. Hinton.
\newblock Imagenet classification with deep convolutional neural networks.
\newblock In \emph{Proceedings of the 25th International Conference on Neural
  Information Processing Systems - Volume 1}, NIPS'12, pp.\  1097–1105, Red
  Hook, NY, USA, 2012. Curran Associates Inc.

\bibitem[Krizhevsky et~al.(2014)Krizhevsky, Nair, and
  Hinton]{krizhevsky2014cifar}
Alex Krizhevsky, Vinod Nair, and Geoffrey Hinton.
\newblock The cifar-10 dataset.
\newblock \emph{online: http://www. cs. toronto. edu/kriz/cifar. html},
  55\penalty0 (5), 2014.

\bibitem[Leclerc et~al.(2021)Leclerc, Salman, Ilyas, Vemprala, Engstrom,
  Vineet, Xiao, Zhang, Santurkar, Yang, Kapoor, and Madry]{leclerc2021three}
Guillaume Leclerc, Hadi Salman, Andrew Ilyas, Sai Vemprala, Logan Engstrom,
  Vibhav Vineet, Kai Xiao, Pengchuan Zhang, Shibani Santurkar, Greg Yang,
  Ashish Kapoor, and Aleksander Madry.
\newblock 3db: A framework for debugging computer vision models.
\newblock In \emph{Arxiv preprint arXiv:2106.03805}, 2021.

\bibitem[Liu et~al.(2015)Liu, Luo, Wang, and Tang]{liu2015faceattributes}
Ziwei Liu, Ping Luo, Xiaogang Wang, and Xiaoou Tang.
\newblock Deep learning face attributes in the wild.
\newblock In \emph{Proceedings of International Conference on Computer Vision
  (ICCV)}, December 2015.

\bibitem[Mottaghi et~al.(2014)Mottaghi, Chen, Liu, Cho, Lee, Fidler, Urtasun,
  and Yuille]{Mottaghi_2014_CVPR}
Roozbeh Mottaghi, Xianjie Chen, Xiaobai Liu, Nam-Gyu Cho, Seong-Whan Lee, Sanja
  Fidler, Raquel Urtasun, and Alan Yuille.
\newblock The role of context for object detection and semantic segmentation in
  the wild.
\newblock In \emph{Proceedings of the IEEE Conference on Computer Vision and
  Pattern Recognition (CVPR)}, June 2014.

\bibitem[Oliva \& Torralba(2007)Oliva and Torralba]{oliva2007role}
Aude Oliva and Antonio Torralba.
\newblock The role of context in object recognition.
\newblock \emph{Trends in cognitive sciences}, 11\penalty0 (12):\penalty0
  520--527, 2007.

\bibitem[Paszke et~al.(2019)Paszke, Gross, Massa, Lerer, Bradbury, Chanan,
  Killeen, Lin, Gimelshein, Antiga, Desmaison, Kopf, Yang, DeVito, Raison,
  Tejani, Chilamkurthy, Steiner, Fang, Bai, and Chintala]{pytorch}
Adam Paszke, Sam Gross, Francisco Massa, Adam Lerer, James Bradbury, Gregory
  Chanan, Trevor Killeen, Zeming Lin, Natalia Gimelshein, Luca Antiga, Alban
  Desmaison, Andreas Kopf, Edward Yang, Zachary DeVito, Martin Raison, Alykhan
  Tejani, Sasank Chilamkurthy, Benoit Steiner, Lu~Fang, Junjie Bai, and Soumith
  Chintala.
\newblock Pytorch: An imperative style, high-performance deep learning library.
\newblock In H.~Wallach, H.~Larochelle, A.~Beygelzimer, F.~d\textquotesingle
  Alch\'{e}-Buc, E.~Fox, and R.~Garnett (eds.), \emph{Advances in Neural
  Information Processing Systems 32}, pp.\  8024--8035. Curran Associates,
  Inc., 2019.

\bibitem[Radford et~al.(2021)Radford, Kim, Hallacy, Ramesh, Goh, Agarwal,
  Sastry, Askell, Mishkin, Clark, et~al.]{radford2021learning}
Alec Radford, Jong~Wook Kim, Chris Hallacy, Aditya Ramesh, Gabriel Goh,
  Sandhini Agarwal, Girish Sastry, Amanda Askell, Pamela Mishkin, Jack Clark,
  et~al.
\newblock Learning transferable visual models from natural language
  supervision.
\newblock \emph{arXiv preprint arXiv:2103.00020}, 2021.

\bibitem[Rosenfeld et~al.(2018)Rosenfeld, Zemel, and
  Tsotsos]{rosenfeld2018elephant}
Amir Rosenfeld, Richard Zemel, and John~K Tsotsos.
\newblock The elephant in the room.
\newblock \emph{arXiv preprint arXiv:1808.03305}, 2018.

\bibitem[Russakovsky et~al.(2015)Russakovsky, Deng, Su, Krause, Satheesh, Ma,
  Huang, Karpathy, Khosla, Bernstein, Berg, and Fei-Fei]{ilsvrc2015}
Olga Russakovsky, Jia Deng, Hao Su, Jonathan Krause, Sanjeev Satheesh, Sean Ma,
  Zhiheng Huang, Andrej Karpathy, Aditya Khosla, Michael~S. Bernstein,
  Alexander~C. Berg, and Li~Fei-Fei.
\newblock Imagenet large scale visual recognition challenge.
\newblock \emph{International Journal of Computer Vision}, 115:\penalty0
  211--252, 2015.

\bibitem[Sagawa* et~al.(2020)Sagawa*, Koh*, Hashimoto, and
  Liang]{Sagawa*2020Distributionally}
Shiori Sagawa*, Pang~Wei Koh*, Tatsunori~B. Hashimoto, and Percy Liang.
\newblock Distributionally robust neural networks.
\newblock In \emph{International Conference on Learning Representations}, 2020.
\newblock URL \url{https://openreview.net/forum?id=ryxGuJrFvS}.

\bibitem[Sastry \& Oore(2020)Sastry and Oore]{pmlr-v119-sastry20a}
Chandramouli~Shama Sastry and Sageev Oore.
\newblock Detecting out-of-distribution examples with {G}ram matrices.
\newblock In Hal~Daumé III and Aarti Singh (eds.), \emph{Proceedings of the
  37th International Conference on Machine Learning}, volume 119 of
  \emph{Proceedings of Machine Learning Research}, pp.\  8491--8501. PMLR,
  13--18 Jul 2020.
\newblock URL \url{https://proceedings.mlr.press/v119/sastry20a.html}.

\bibitem[Sauer \& Geiger(2021)Sauer and Geiger]{sauer2021counterfactual}
Axel Sauer and Andreas Geiger.
\newblock Counterfactual generative networks.
\newblock In \emph{International Conference on Learning Representations}, 2021.
\newblock URL \url{https://openreview.net/forum?id=BXewfAYMmJw}.

\bibitem[Selvaraju et~al.(2017)Selvaraju, Cogswell, Das, Vedantam, Parikh, and
  Batra]{selvaraju2017grad}
Ramprasaath~R Selvaraju, Michael Cogswell, Abhishek Das, Ramakrishna Vedantam,
  Devi Parikh, and Dhruv Batra.
\newblock Grad-cam: Visual explanations from deep networks via gradient-based
  localization.
\newblock In \emph{Proceedings of the IEEE international conference on computer
  vision}, pp.\  618--626, 2017.

\bibitem[Singla et~al.(2021)Singla, Nushi, Shah, Kamar, and
  Horvitz]{singla2021understanding}
Sahil Singla, Besmira Nushi, Shital Shah, Ece Kamar, and Eric Horvitz.
\newblock Understanding failures of deep networks via robust feature
  extraction.
\newblock In \emph{Proceedings of the IEEE/CVF Conference on Computer Vision
  and Pattern Recognition}, pp.\  12853--12862, 2021.

\bibitem[Tan \& Le(2019)Tan and Le]{tan2019efficientnet}
Mingxing Tan and Quoc Le.
\newblock Efficientnet: Rethinking model scaling for convolutional neural
  networks.
\newblock In \emph{International Conference on Machine Learning}, pp.\
  6105--6114. PMLR, 2019.

\bibitem[Wightman(2019)]{rw2019timm}
Ross Wightman.
\newblock Pytorch image models.
\newblock \url{https://github.com/rwightman/pytorch-image-models}, 2019.

\bibitem[Wong et~al.(2021)Wong, Santurkar, and Madry]{sparsewong21b}
Eric Wong, Shibani Santurkar, and Aleksander Madry.
\newblock Leveraging sparse linear layers for debuggable deep networks.
\newblock In Marina Meila and Tong Zhang (eds.), \emph{Proceedings of the 38th
  International Conference on Machine Learning}, volume 139 of
  \emph{Proceedings of Machine Learning Research}, pp.\  11205--11216. PMLR,
  18--24 Jul 2021.
\newblock URL \url{https://proceedings.mlr.press/v139/wong21b.html}.

\bibitem[Xiao et~al.(2020)Xiao, Engstrom, Ilyas, and Madry]{xiao2020noise}
Kai Xiao, Logan Engstrom, Andrew Ilyas, and Aleksander Madry.
\newblock Noise or signal: The role of image backgrounds in object recognition.
\newblock \emph{ArXiv preprint arXiv:2006.09994}, 2020.

\bibitem[Xie et~al.(2017)Xie, Girshick, Doll{\'a}r, Tu, and
  He]{xie2017aggregated}
Saining Xie, Ross Girshick, Piotr Doll{\'a}r, Zhuowen Tu, and Kaiming He.
\newblock Aggregated residual transformations for deep neural networks.
\newblock In \emph{Proceedings of the IEEE conference on computer vision and
  pattern recognition}, pp.\  1492--1500, 2017.

\bibitem[Yu et~al.(2020)Yu, Chen, Wang, Xian, Chen, Liu, Madhavan, and
  Darrell]{yu2020bdd100k}
Fisher Yu, Haofeng Chen, Xin Wang, Wenqi Xian, Yingying Chen, Fangchen Liu,
  Vashisht Madhavan, and Trevor Darrell.
\newblock Bdd100k: A diverse driving dataset for heterogeneous multitask
  learning.
\newblock In \emph{Proceedings of the IEEE/CVF conference on computer vision
  and pattern recognition}, pp.\  2636--2645, 2020.

\bibitem[Zagoruyko \& Komodakis(2016)Zagoruyko and
  Komodakis]{zagoruyko2016wide}
Sergey Zagoruyko and Nikos Komodakis.
\newblock Wide residual networks.
\newblock \emph{arXiv preprint arXiv:1605.07146}, 2016.

\end{thebibliography}
\bibliographystyle{iclr2022_conference}

%\section*{A.1 Appendix}

\clearpage

\noindent {\bf \LARGE Appendix}

\appendix

% \clearpage
\section{Image Search}\label{app:image-search}
The images for our dataset were collected using queries formed as a concatenation of an object label and one of the phrases from below:

\begin{table}[ht]
\begin{tabular}{c c}
Attribute & Phrases used in queries \\
\hline
\textit{``raining''} & ``in rain'' \\
\textit{``foggy''} & ``in fog'' \\
\textit{``snow''} & ``on snow'' \\
\textit{``sand''} & ``in a desert'', ``on sand'' \\
\textit{``forest''} & ``in forest''\\
\textit{``water''} & ``on water''\\
\textit{``night''} & ``at night''\\
\textit{``grass''} & ``on grass''\\
\textit{``street''} & ``on a street'', ``on a road'' \\
\end{tabular}
\end{table}

In addition, we use some class specific queries: \textit{``ship on ice''}, \textit{``ship on a dock''}, \textit{``dog on a couch''}, \textit{``dog on a bed''}, \textit{``dog on the floor''}, \textit{``cat on a couch''}, \textit{``cat on a bed''}, \textit{``cat on the floor''},  \textit{``horse in a stable''}, \textit{``car in a garage''}, \textit{``truck in a garage''}, \textit{``plane in a hangar''}. 

Figure~\ref{fig:more-uncommon} shows some more uncommon images in FOCUS.

\begin{figure*}
\centering
\input{uncommon}
\caption{Some uncommon images in the FOCUS dataset.}\label{fig:more-uncommon}
\end{figure*}

\clearpage

\begin{figure*}[t]
\centering
\includegraphics[width=\linewidth]{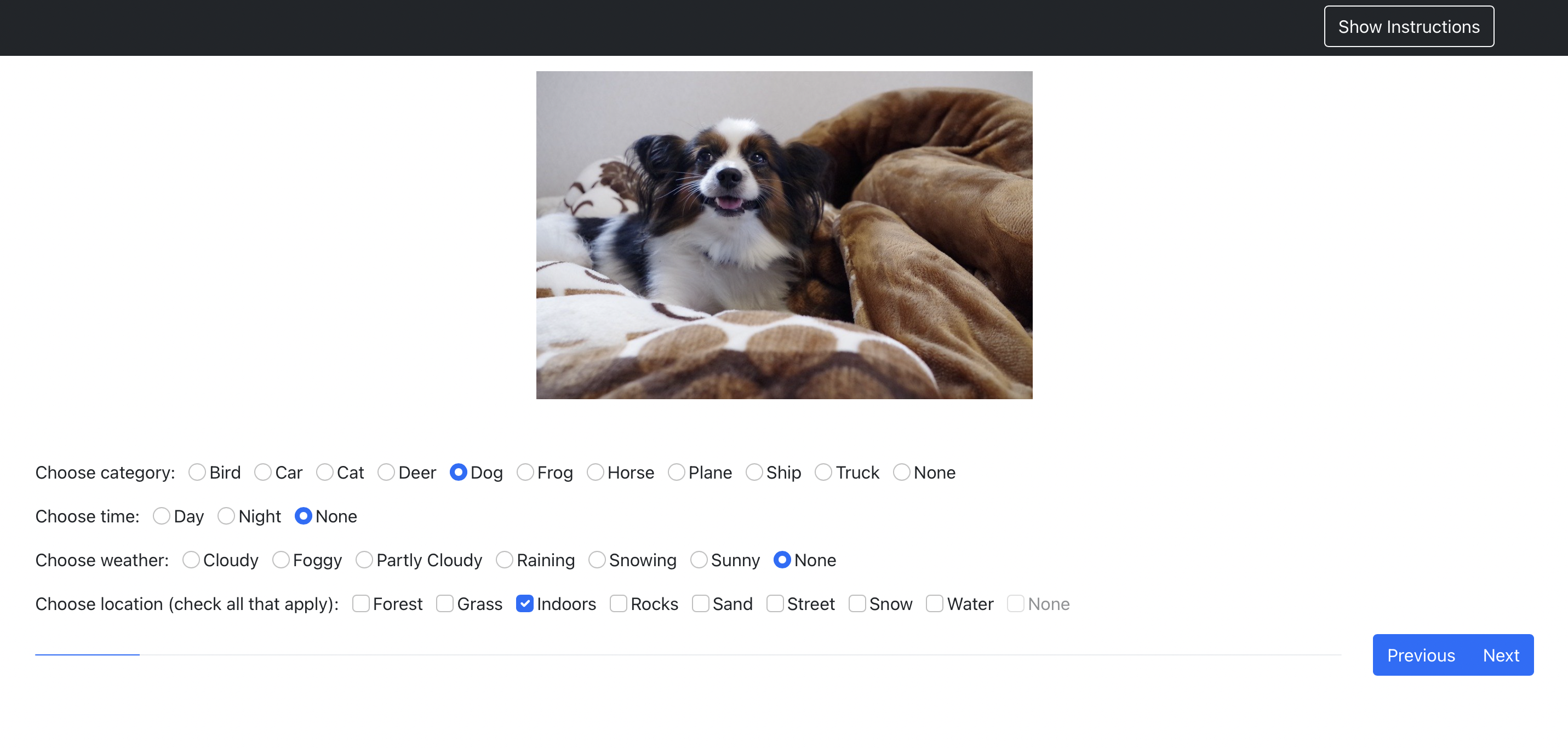}
\caption{Our UI for annotating images in FOCUS.}
\label{fig:hit-ui}
\end{figure*}
\section{ Human Intelligence Tasks (HITs)}\label{app:hit-study}

To gather high quality annotations, we first vet the workers through a qualification process; workers are shown a series of 10 images (in the UI shown in figure~\ref{fig:hit-ui}) from our dataset for which the ground truth is known (these images were annotated by us manually). We qualify workers who have done well on this qualification test. Worker's annotations were checked manually in this stage instead of using a strict threshold as there is an element of subjectivity to the annotations. In the second stage, our HITs have 25 images each; 23 of which are unannotated and 2 are from the subset that were annotated by us. We use these 2 images as a way to track workers' annotation accuracy. Each image is annotated by two workers and we pick annotations of the worker who has the higher annotation accuracy on the 2 ``check'' images in that HIT. Workers received a base pay of \$0.67 per HIT (25 images) which takes an average of around 5 minutes to annotate. A bonus of \$30 was paid for completion of 100 HITs. This payment structure has created an incentive for workers to annotate a large number of images.

\clearpage

\section{Accuracy as a function of class and attributes}\label{app:acc-vs-att}

This section shows the accuracy of different models for each class and attribute. Uncommon attributes are highlighted in shades of orange, while common attributes are highlighted in shades of blue. The last row (except the bottom right-most value) shows the overall accuracy for the corresponding attribute. Additionally, the last column is the generalization gap with respect to the corresponding attribute. The first 10 values in the last column are class specific generalization gaps, while the last (i.e., bottom rightmost) value is the aggregate generalization gap defined in Equation~\ref{eq:generalization_gap} (and reported in table~\ref{tab:gen-gaps}).

An image can have a combination of common and uncommon attributes (e.g., \textit{``cat on street at night''} --- uncommon time but common location). Such images may appear in a ``blue'' cell in one of the tables while in a different table they may appear in ``orange''. This explains why for some classes the models seem to do better at ``night'' than in day. Note that majority of the uncommon weather conditions: \textit{(``raining'', ``snowing'', ``foggy'')} occur during the day and they are potentially decreasing the classification accuracy so much that it outweighs the drop due to ``night'' and leads to this apparently paradoxical result.

\begin{figure*}[ht]
\begin{subfigure}{0.49\textwidth}
\includegraphics[width=\linewidth]{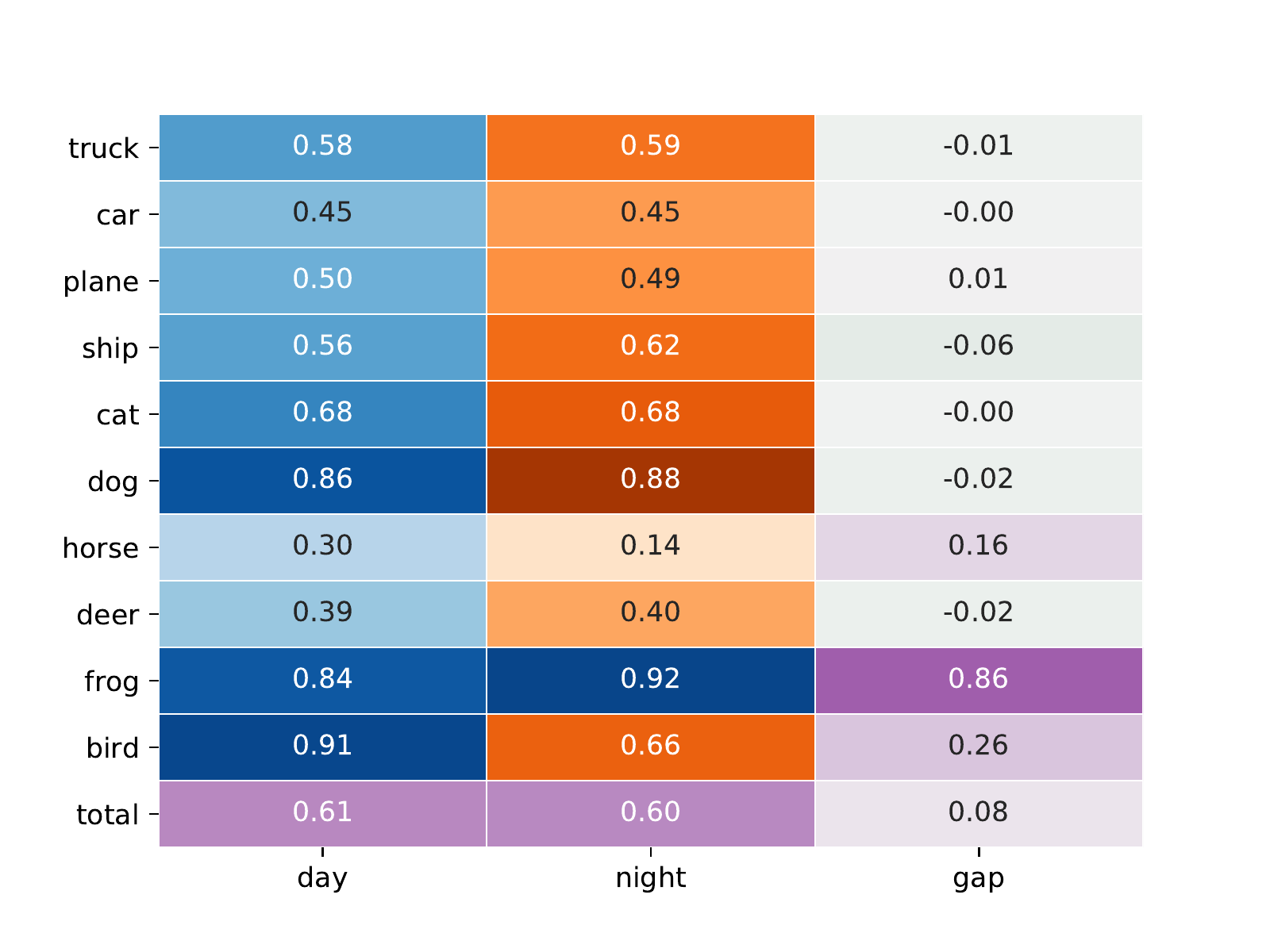}
\caption{Category vs Time of Day}
\end{subfigure}
\begin{subfigure}{0.49\textwidth}
\includegraphics[width=\linewidth]{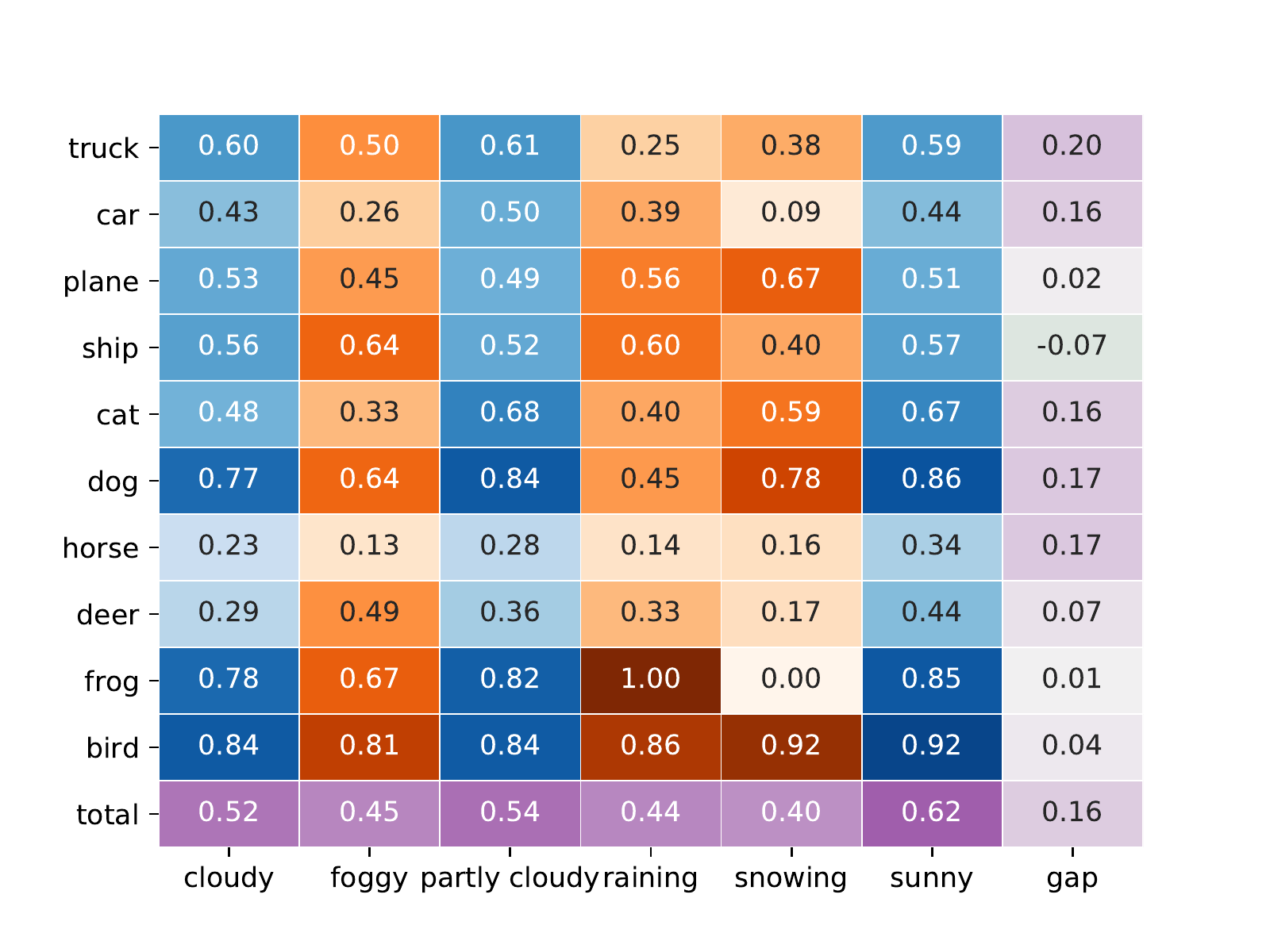}
\caption{Category vs Weather}
\end{subfigure}

\begin{subfigure}{\textwidth}
\includegraphics[width=\linewidth]{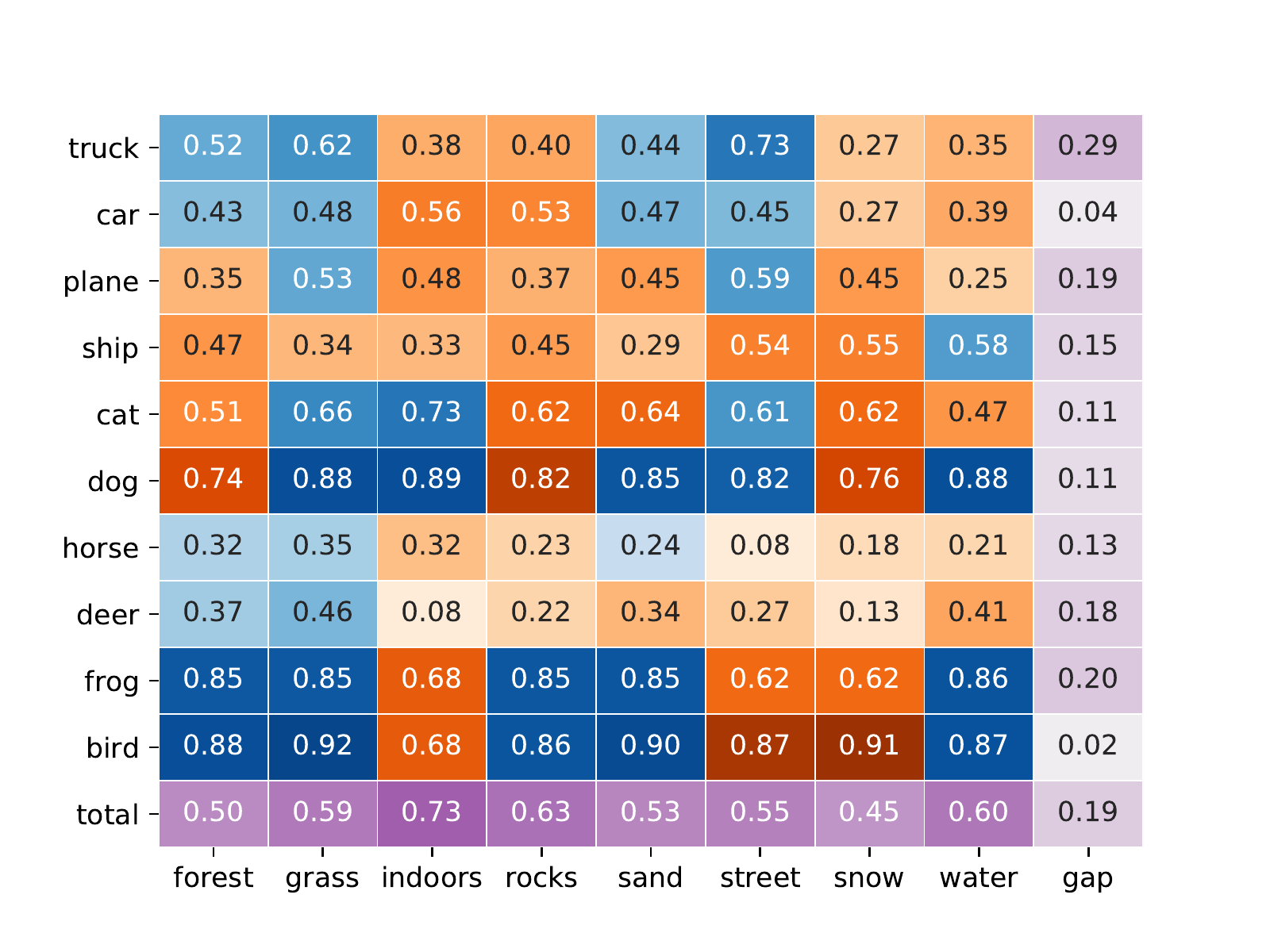}
\caption{Category vs Location}
\end{subfigure}

\caption{Accuracy of ResNet50 for all combinations of classes and attributes.}

\end{figure*}

\newpage

\begin{figure*}[ht]
\begin{subfigure}{0.49\textwidth}
\includegraphics[width=\linewidth]{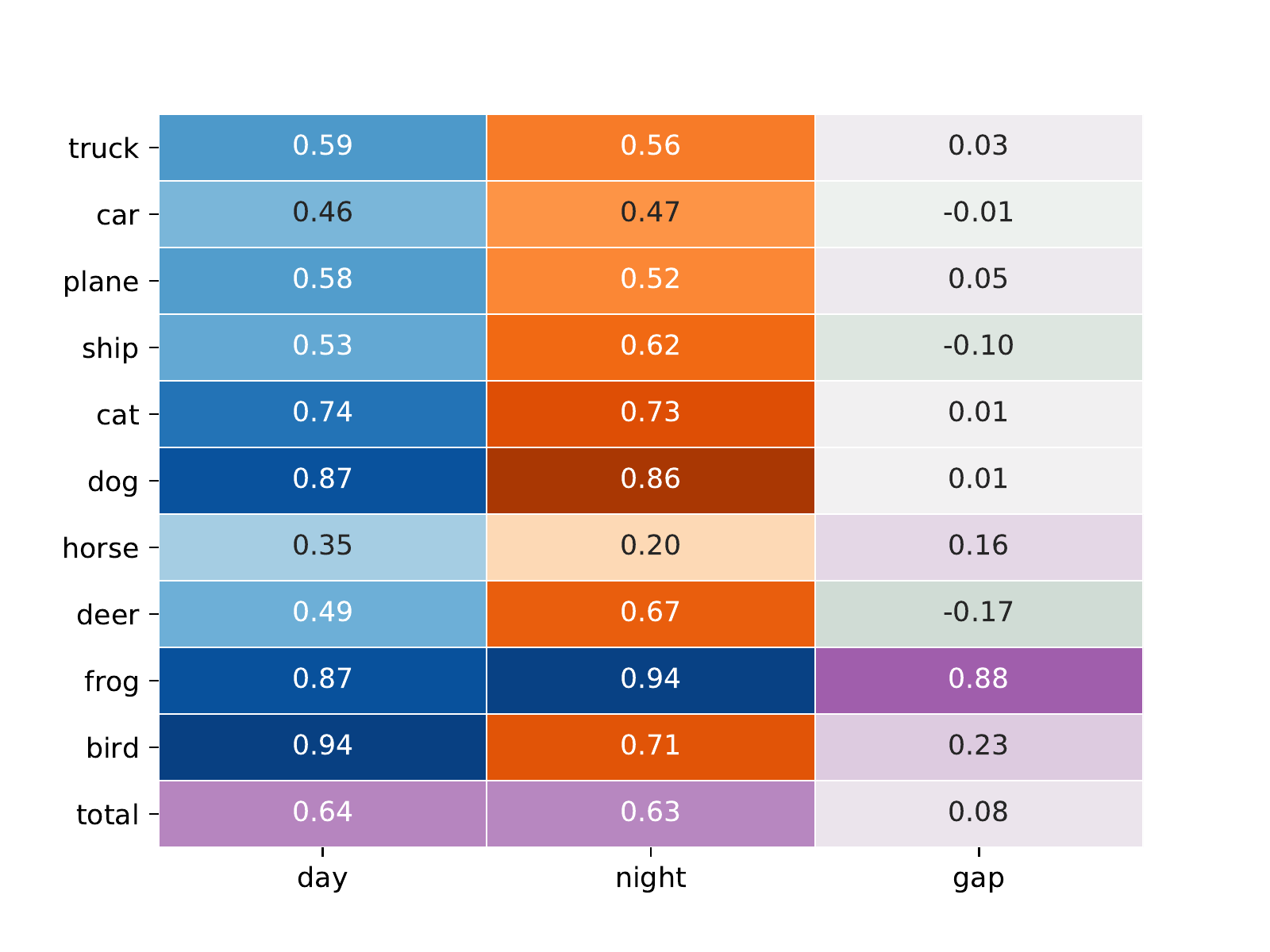}
\caption{Category vs Time of Day}
\end{subfigure}
\begin{subfigure}{0.49\textwidth}
\includegraphics[width=\linewidth]{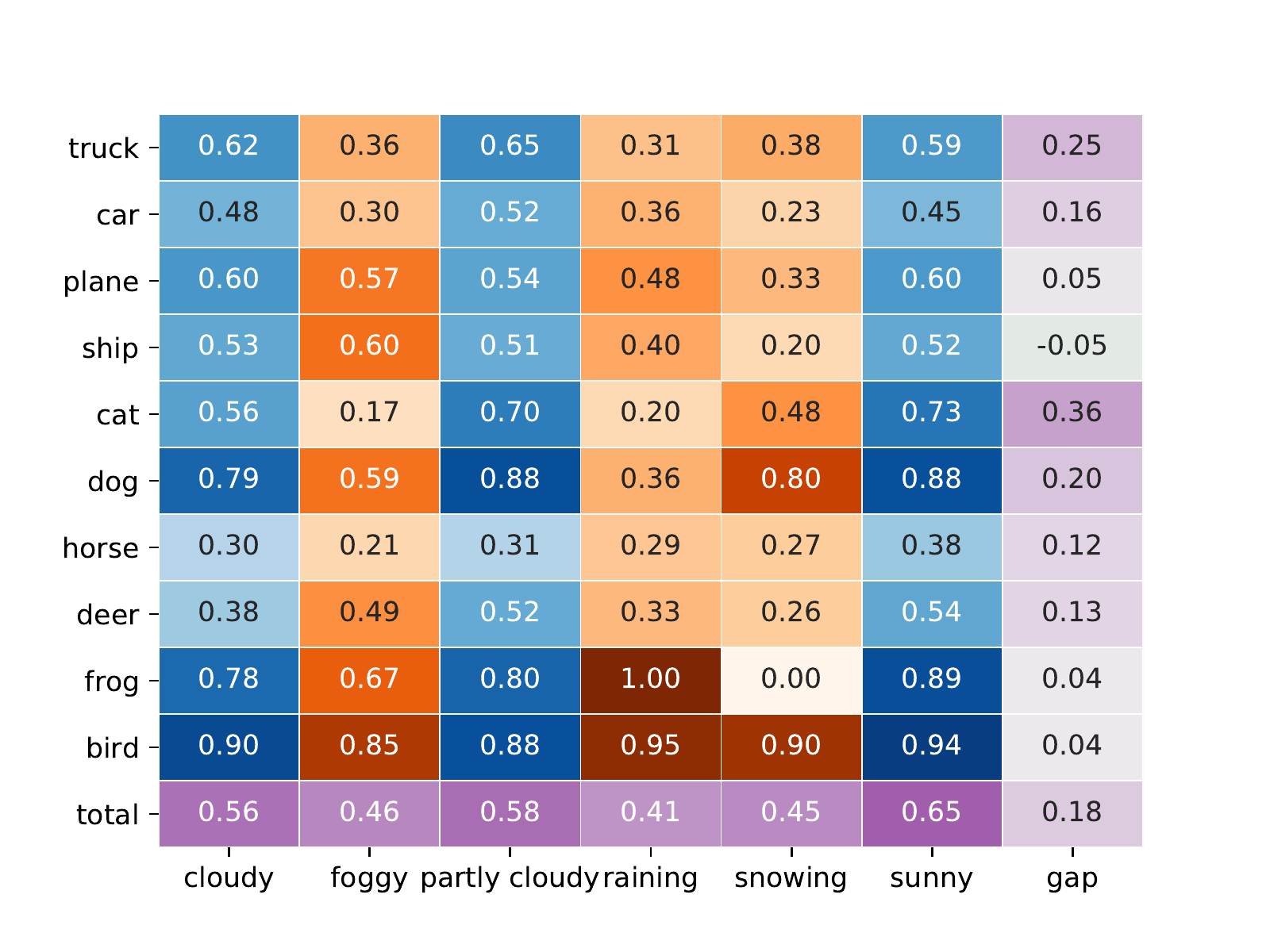}
\caption{Category vs Weather}
\end{subfigure}

\begin{subfigure}{\textwidth}
\includegraphics[width=\linewidth]{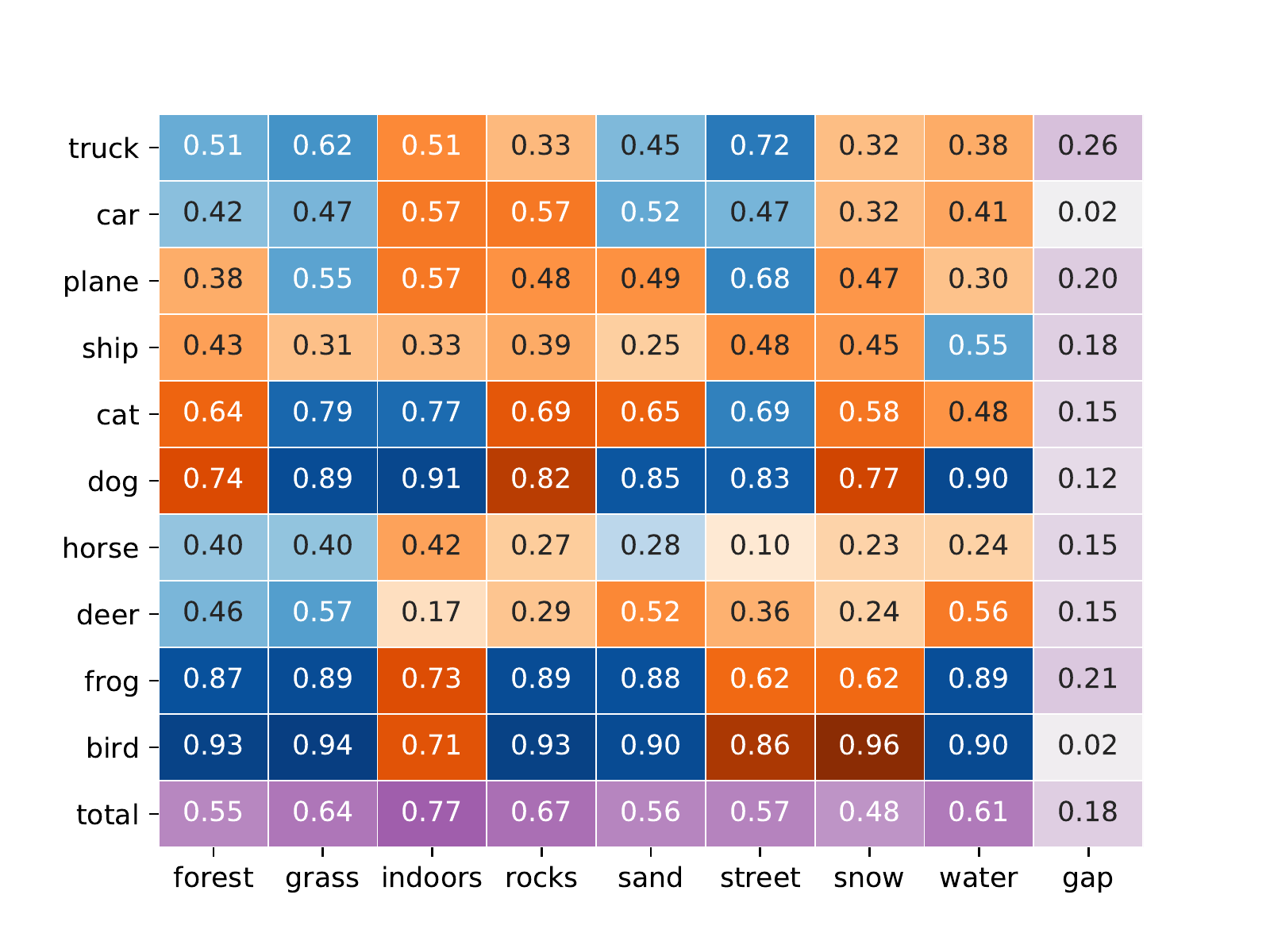}
\caption{Category vs Location}
\end{subfigure}

\caption{Accuracy of Wide-ResNet50-2 for all combinations of classes and attributes.}

\end{figure*}

\newpage

\begin{figure*}[ht]
\begin{subfigure}{0.49\textwidth}
\includegraphics[width=\linewidth]{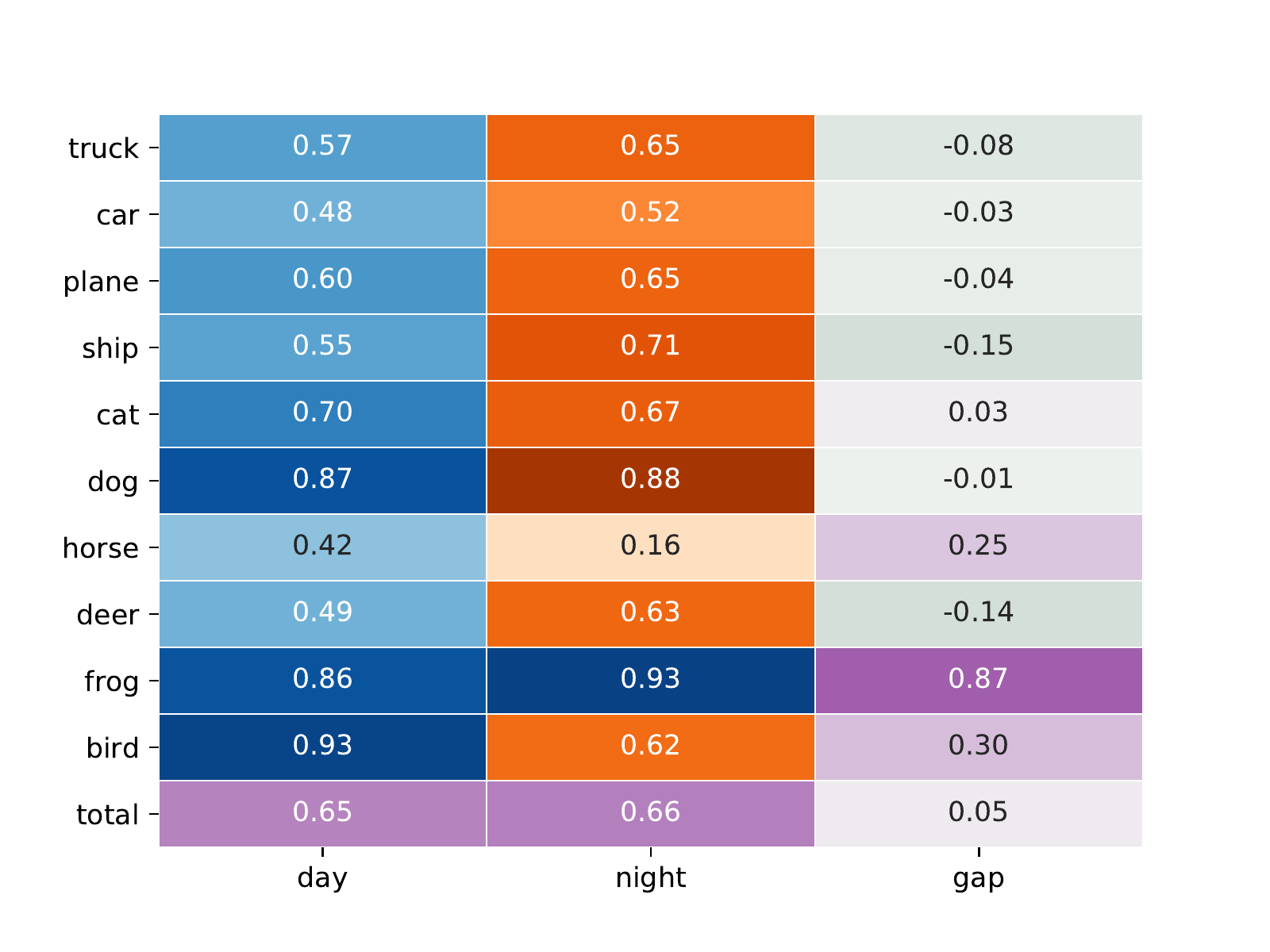}
\caption{Category vs Time of Day}
\end{subfigure}
\begin{subfigure}{0.49\textwidth}
\includegraphics[width=\linewidth]{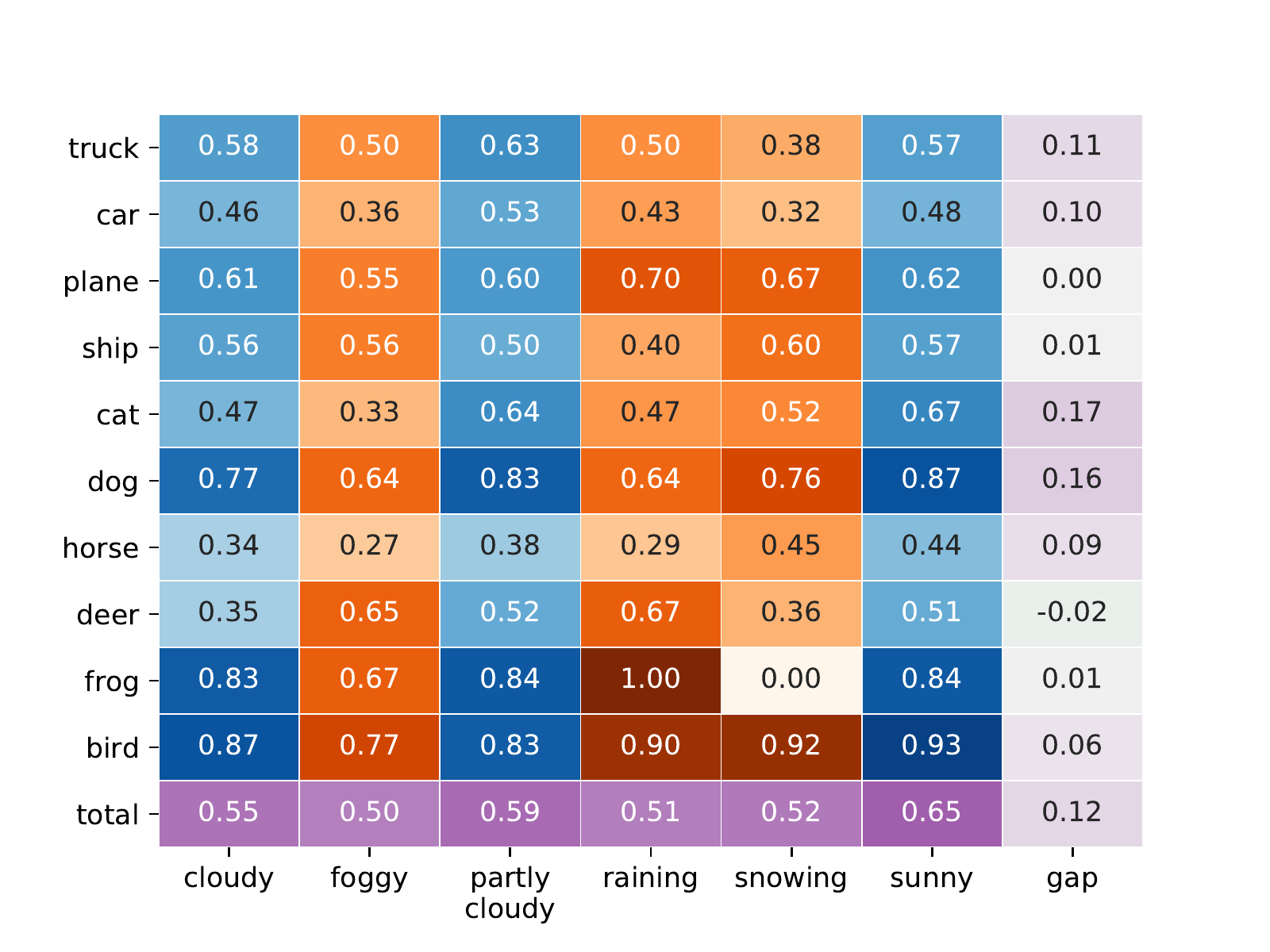}
\caption{Category vs Weather}
\end{subfigure}

\begin{subfigure}{\textwidth}
\includegraphics[width=\linewidth]{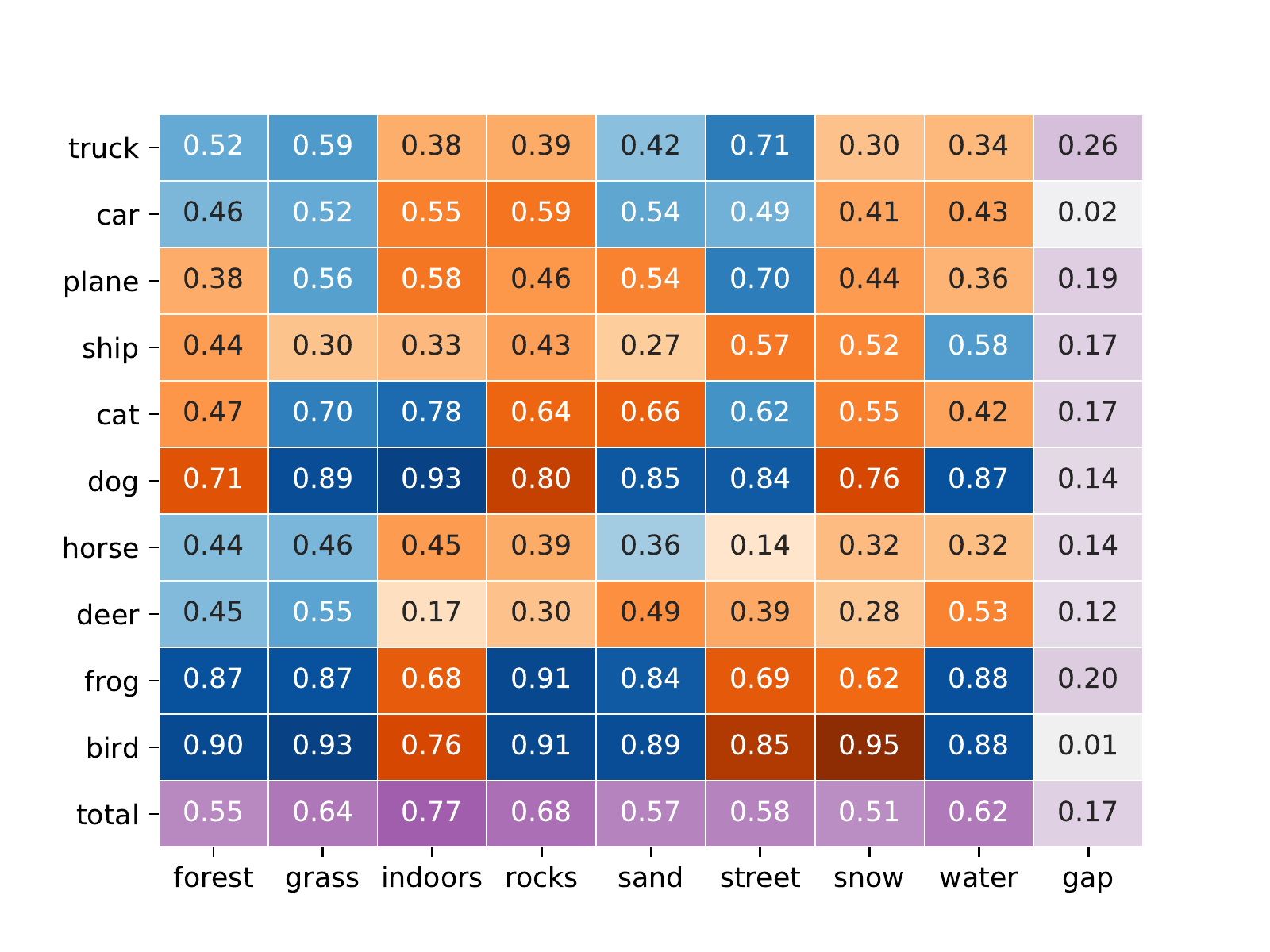}
\caption{Category vs Location}
\end{subfigure}

\caption{Accuracy of EfficientNet-b4 for all combinations of classes and attributes.}

\end{figure*}

\newpage

\begin{figure*}[ht]
\begin{subfigure}{0.49\textwidth}
\includegraphics[width=\linewidth]{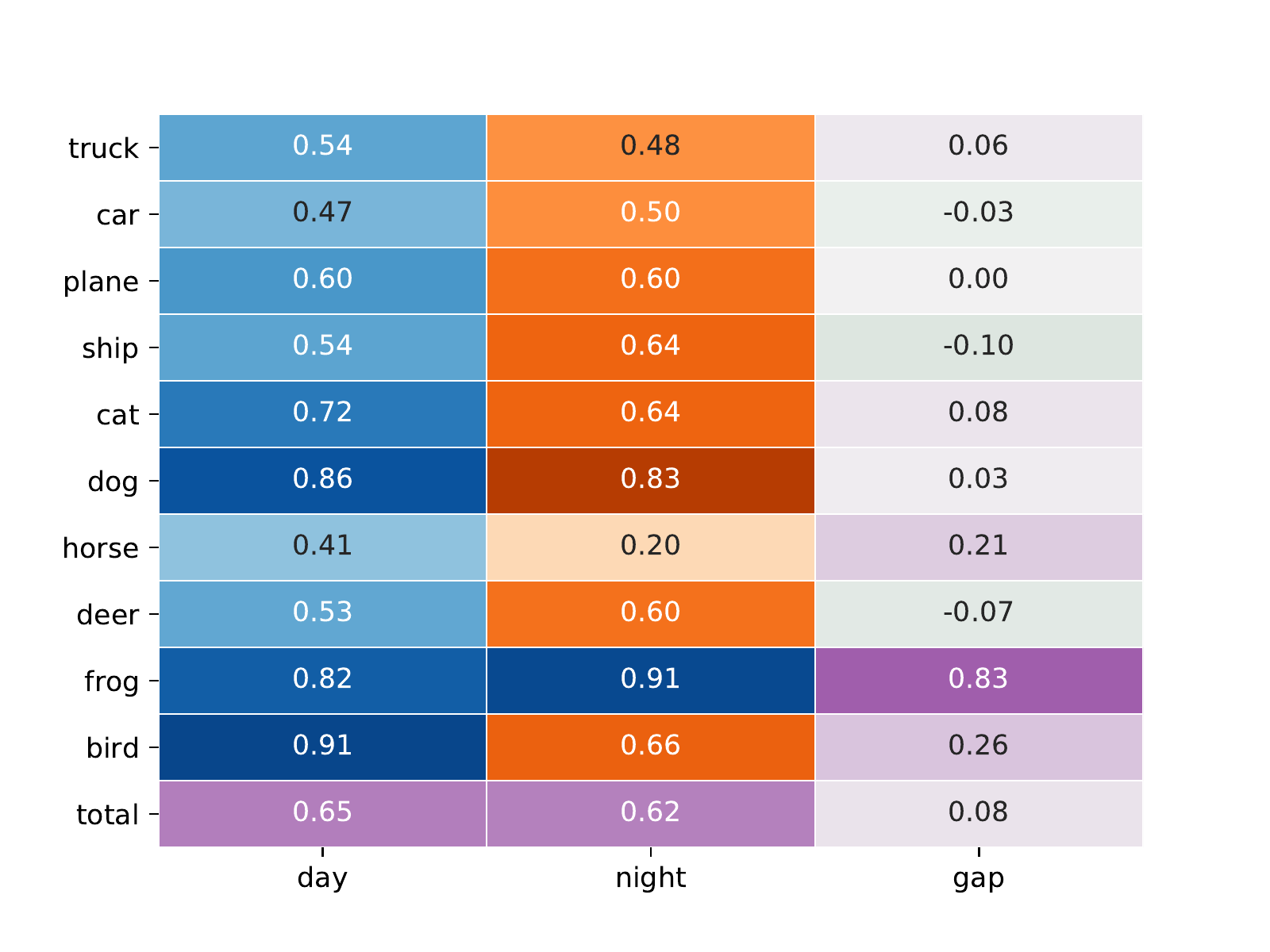}
\caption{Category vs Time of Day}
\end{subfigure}
\begin{subfigure}{0.49\textwidth}
\includegraphics[width=\linewidth]{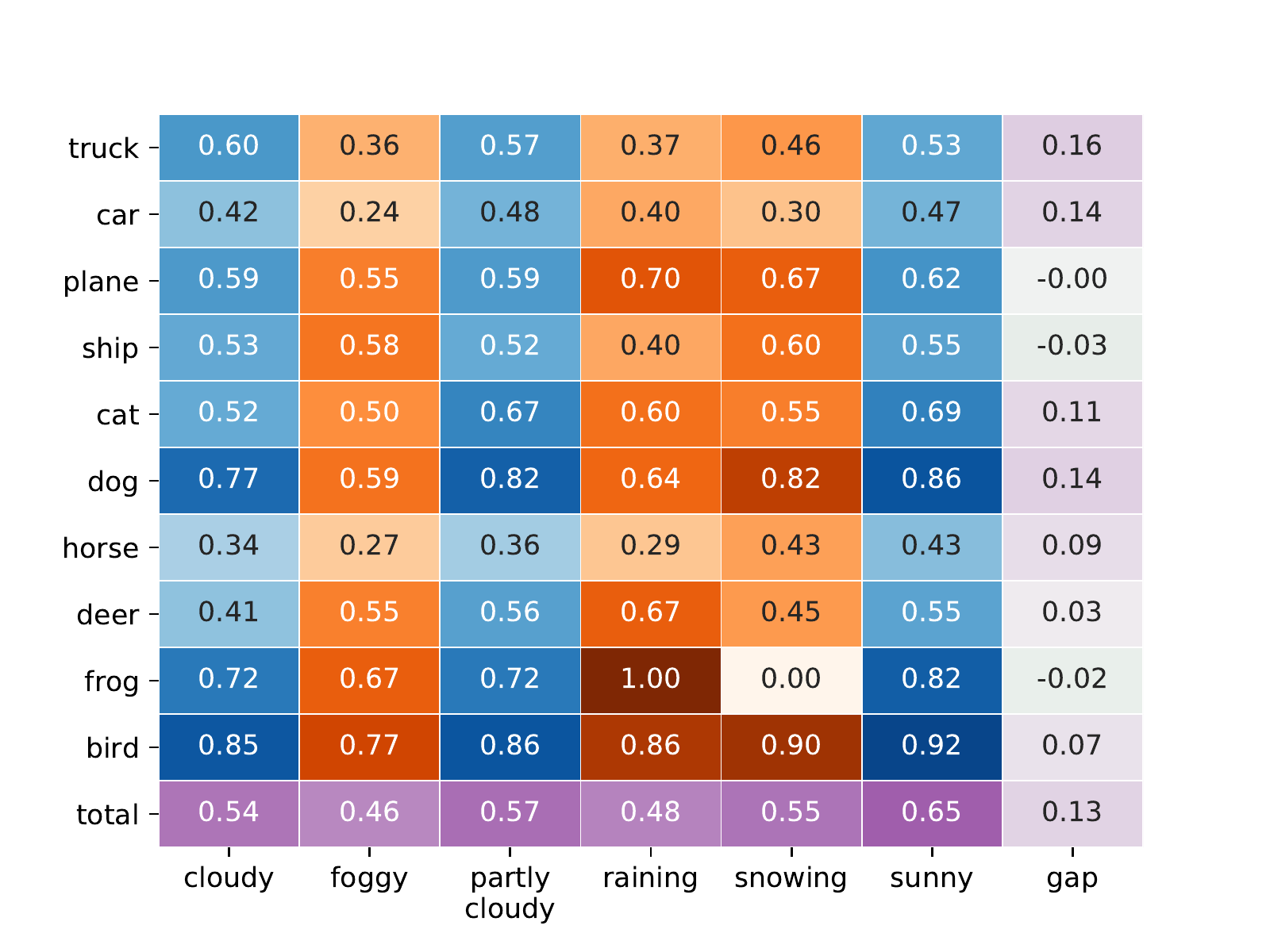}
\caption{Category vs Weather}
\end{subfigure}

\begin{subfigure}{\textwidth}
\includegraphics[width=\linewidth]{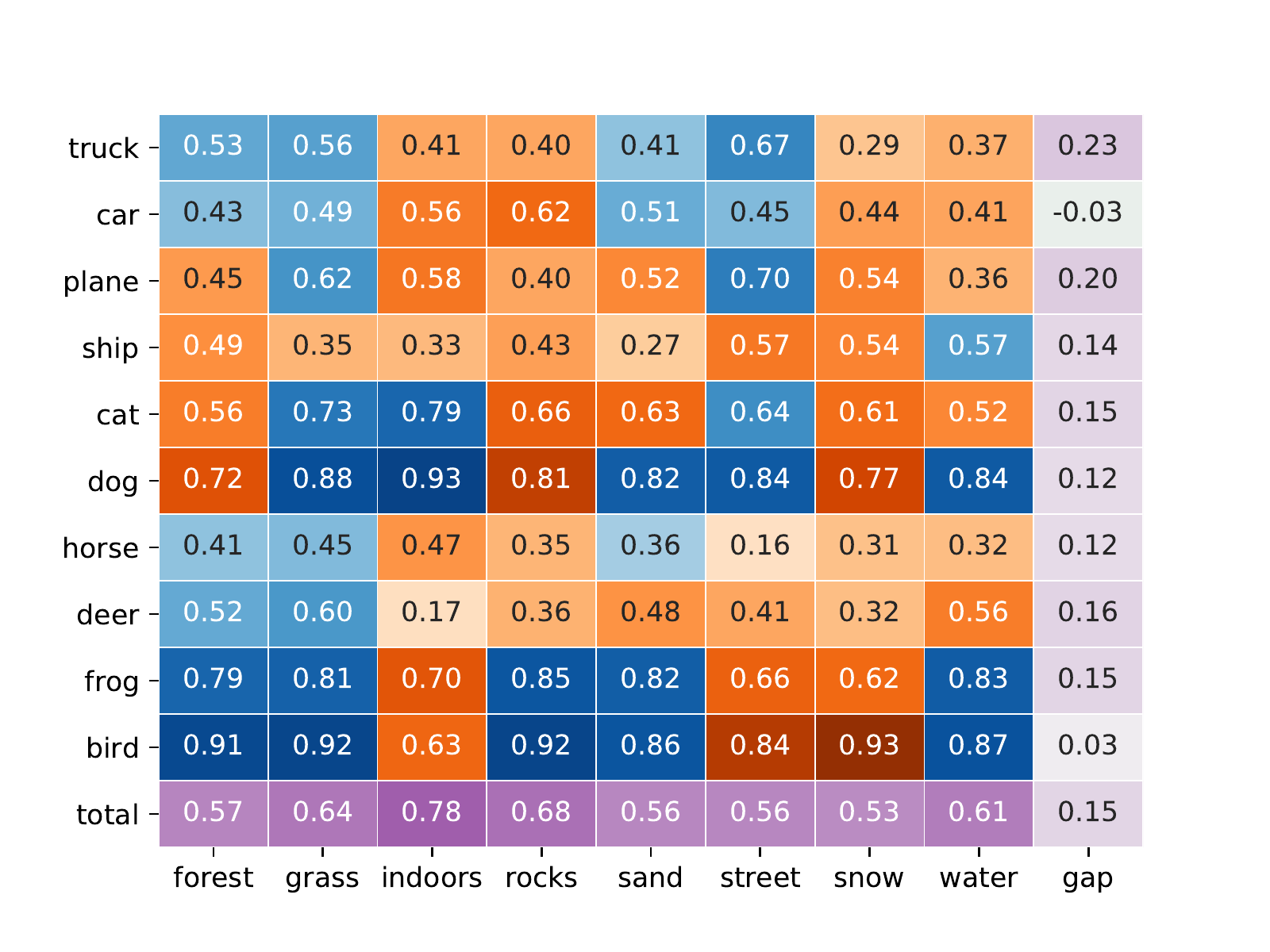}
\caption{Category vs Location}
\end{subfigure}

\caption{Accuracy of EfficientNet-b7 for all combinations of classes and attributes.}

\end{figure*}

\newpage

\begin{figure*}[ht]
\begin{subfigure}{0.49\textwidth}
\includegraphics[width=\linewidth]{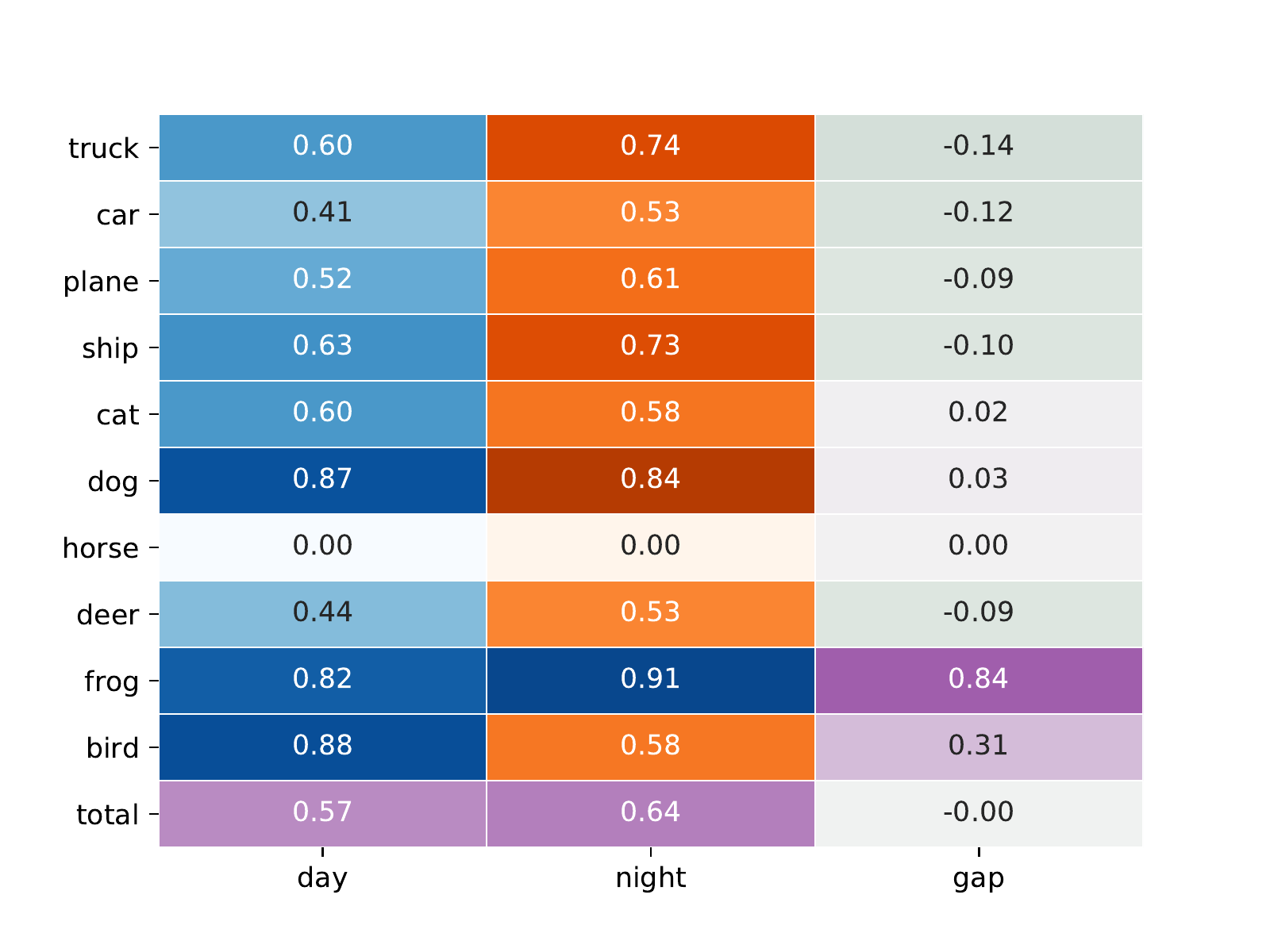}
\caption{Category vs Time of Day}
\end{subfigure}
\begin{subfigure}{0.49\textwidth}
\includegraphics[width=\linewidth]{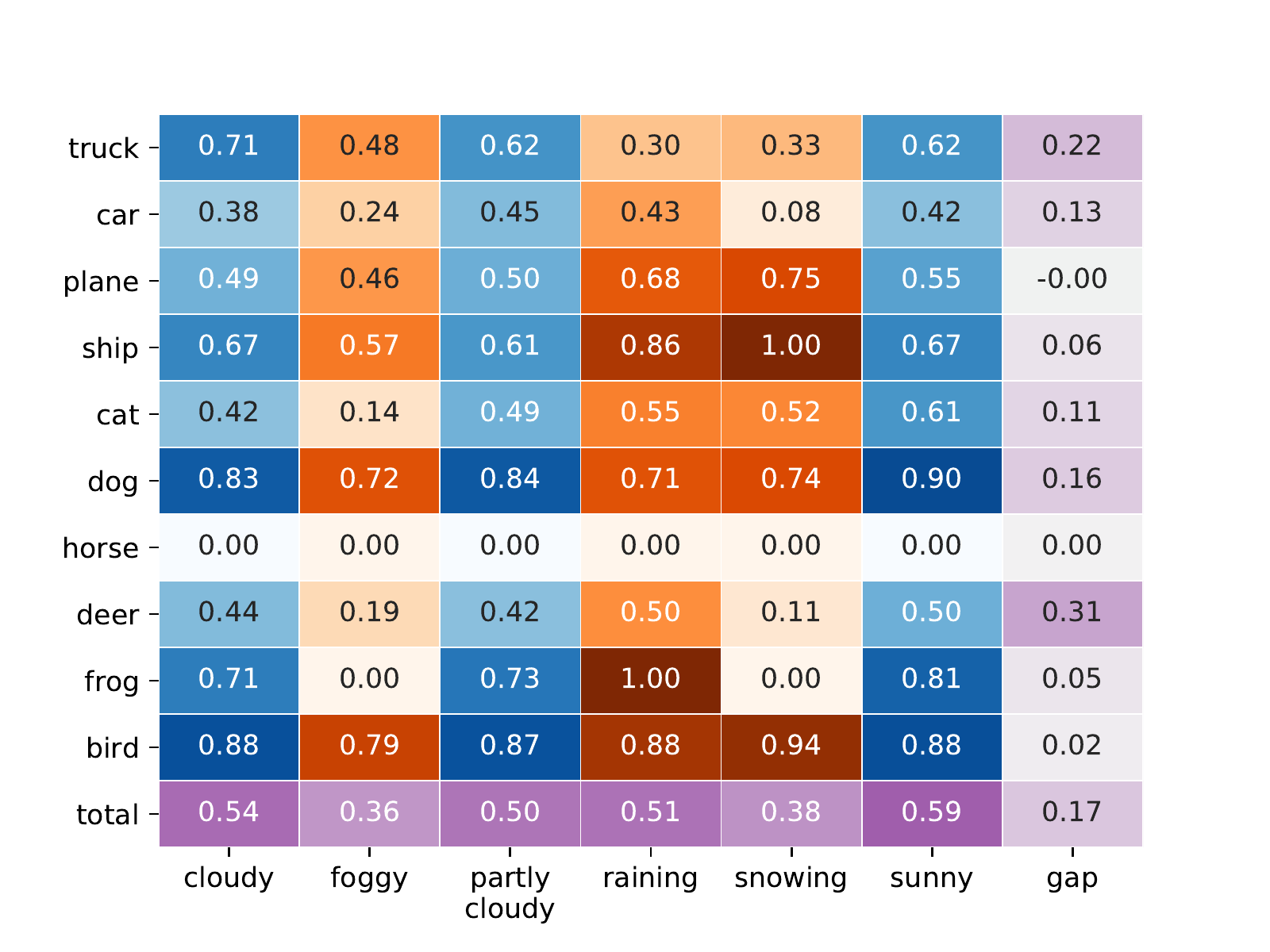}
\caption{Category vs Weather}
\end{subfigure}

\begin{subfigure}{\textwidth}
\includegraphics[width=\linewidth]{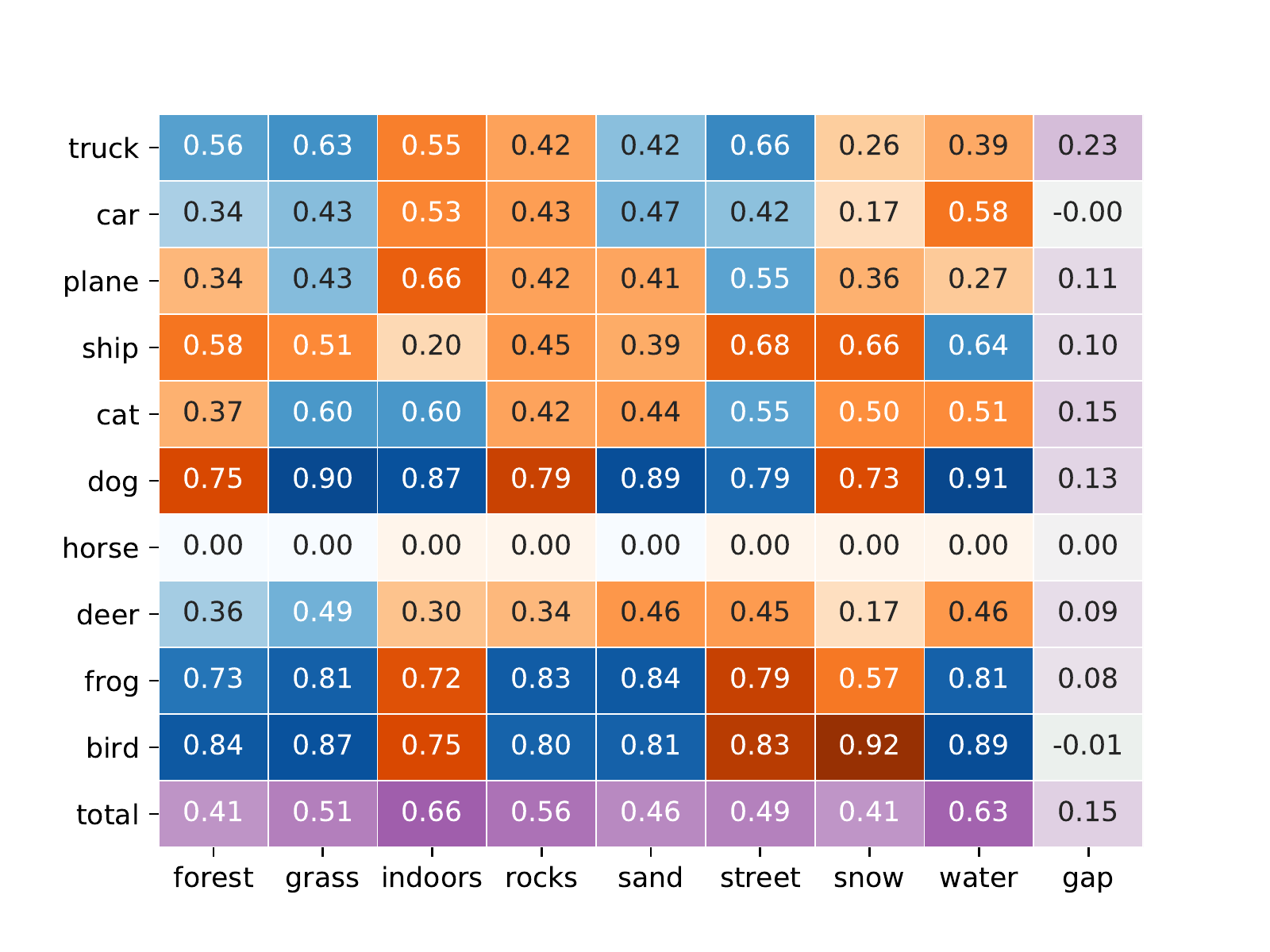}
\caption{Category vs Location}
\end{subfigure}

\caption{Accuracy of CLIP for all combinations of classes and attributes.}

\end{figure*}

\newpage

\begin{figure*}[ht]
\begin{subfigure}{0.49\textwidth}
\includegraphics[width=\linewidth]{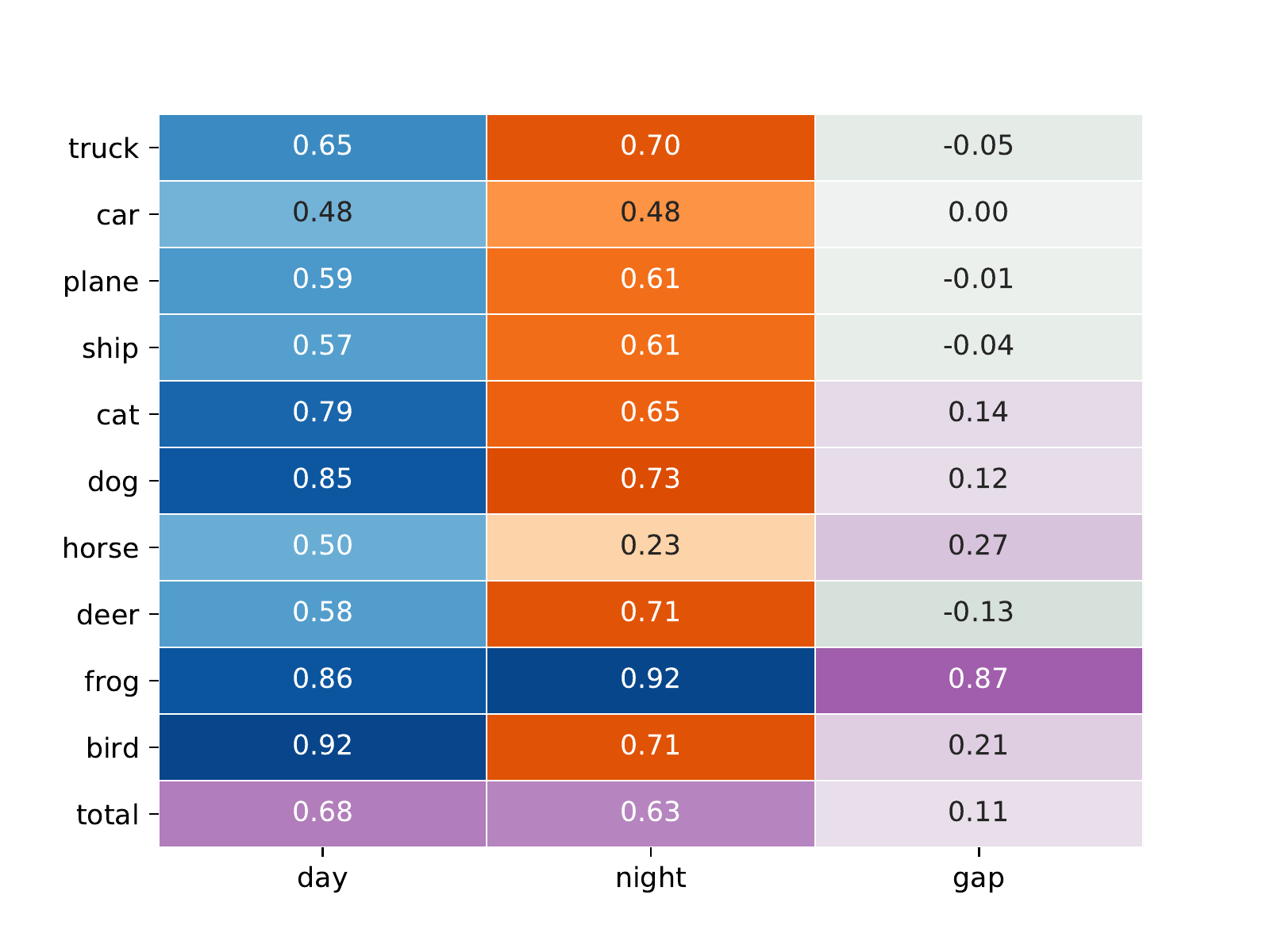}
\caption{Category vs Time of Day}
\end{subfigure}
\begin{subfigure}{0.49\textwidth}
\includegraphics[width=\linewidth]{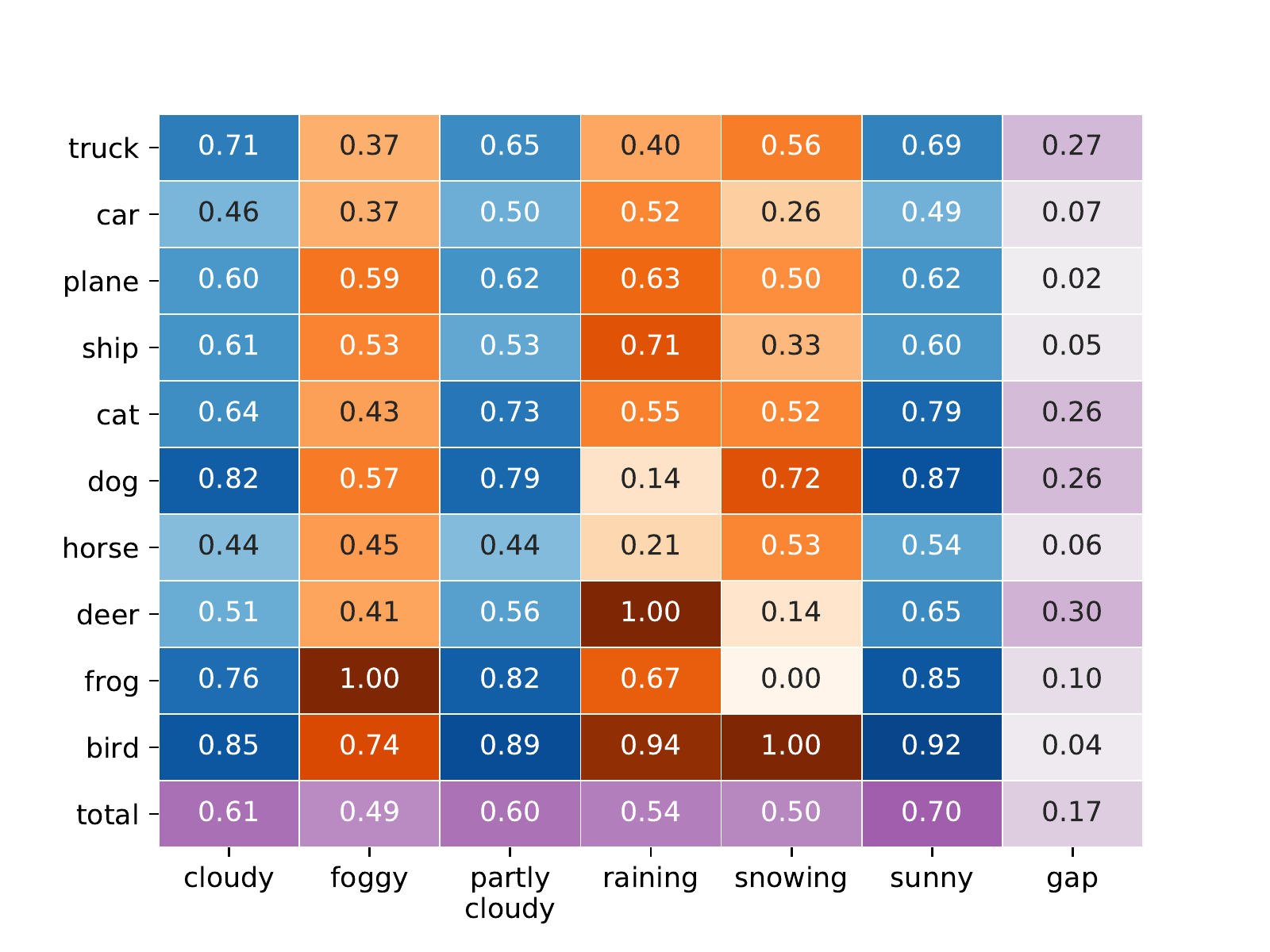}
\caption{Category vs Weather}
\end{subfigure}

\begin{subfigure}{\textwidth}
\includegraphics[width=\linewidth]{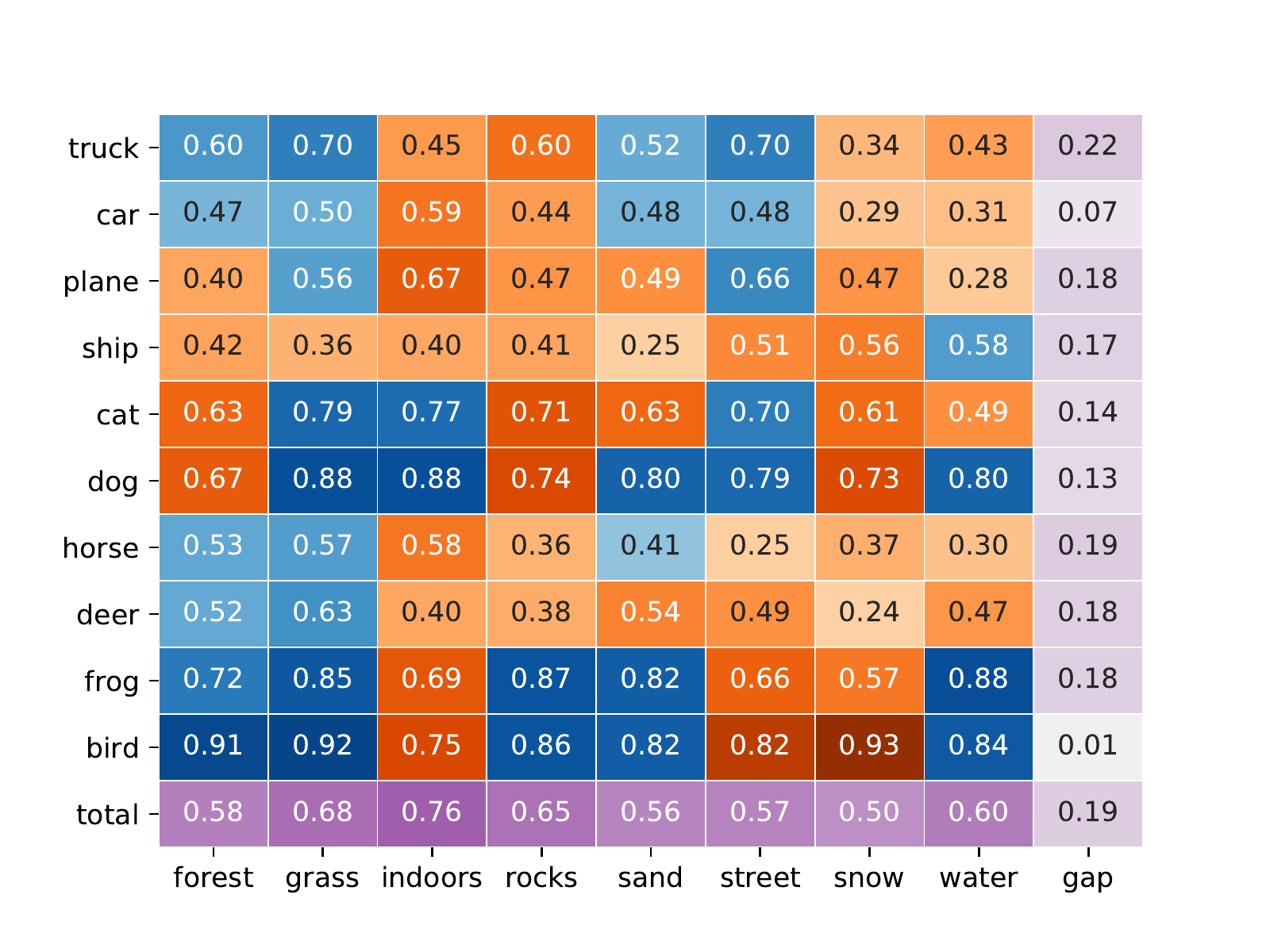}
\caption{Category vs Location}
\end{subfigure}

\caption{Accuracy of ViT-B/16 for all combinations of classes and attributes.}

\end{figure*}

\newpage

\begin{figure*}[ht]
\begin{subfigure}{0.49\textwidth}
\includegraphics[width=\linewidth]{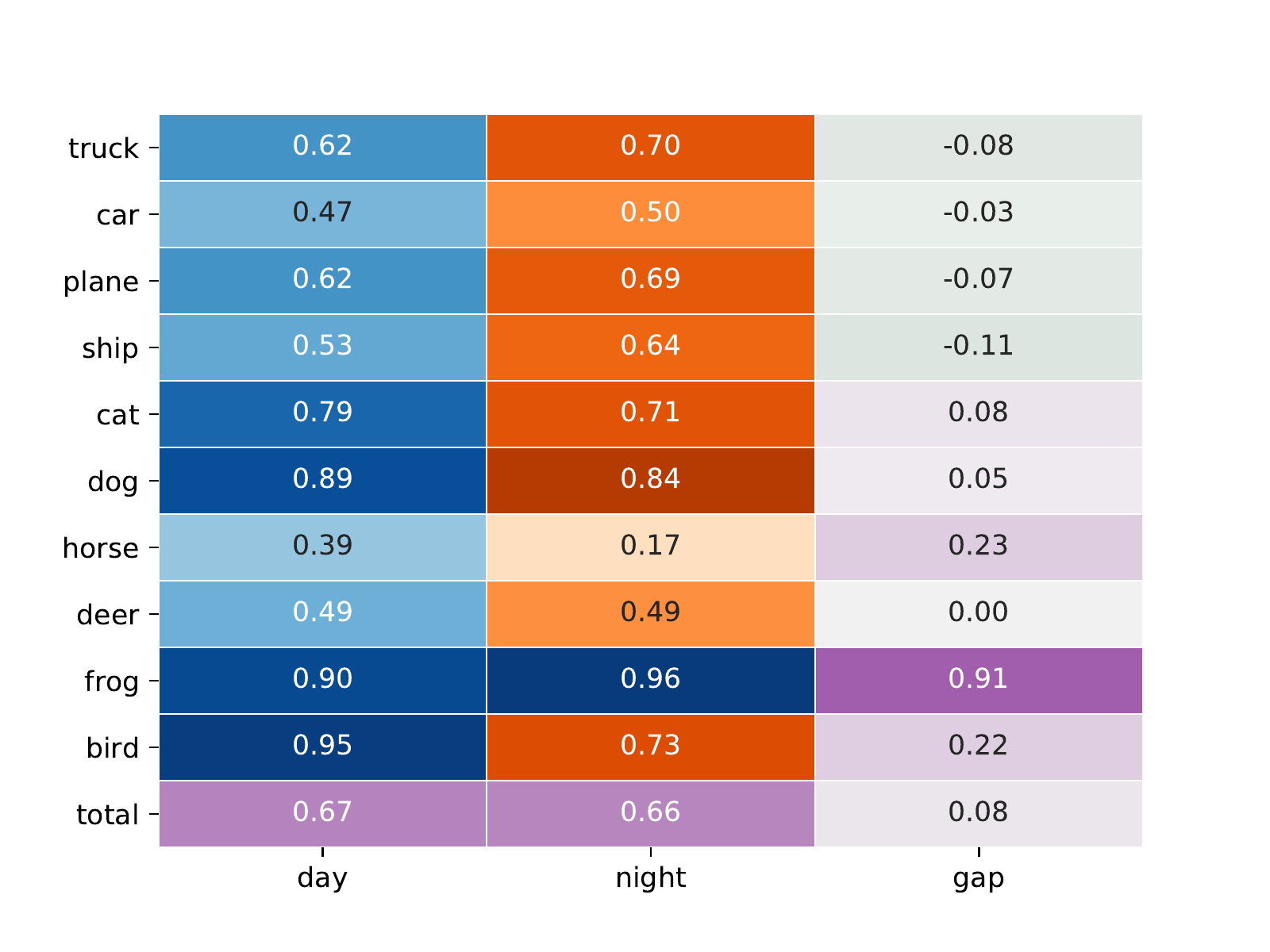}
\caption{Category vs Time of Day}
\end{subfigure}
\begin{subfigure}{0.49\textwidth}
\includegraphics[width=\linewidth]{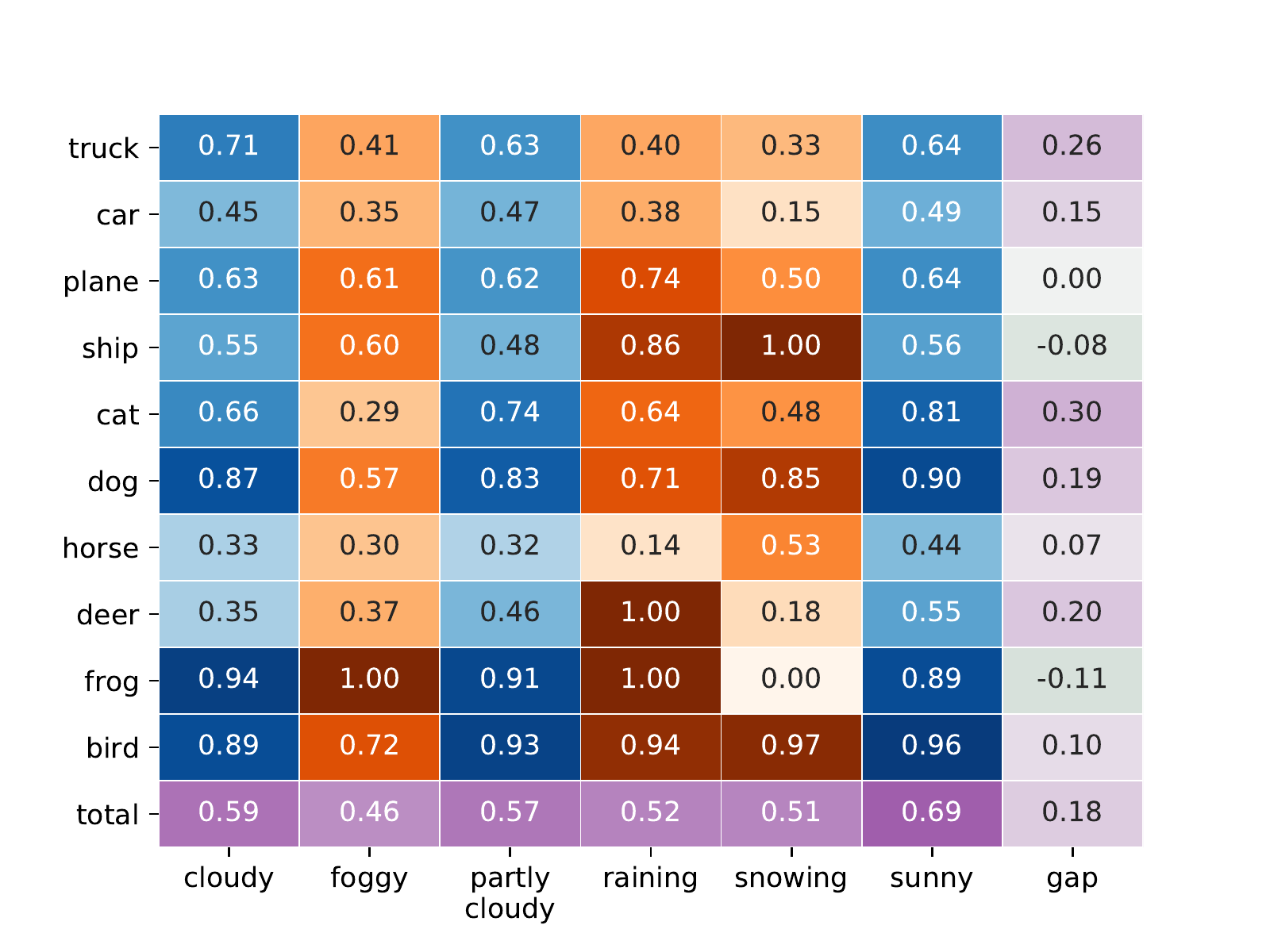}
\caption{Category vs Weather}
\end{subfigure}

\begin{subfigure}{\textwidth}
\includegraphics[width=\linewidth]{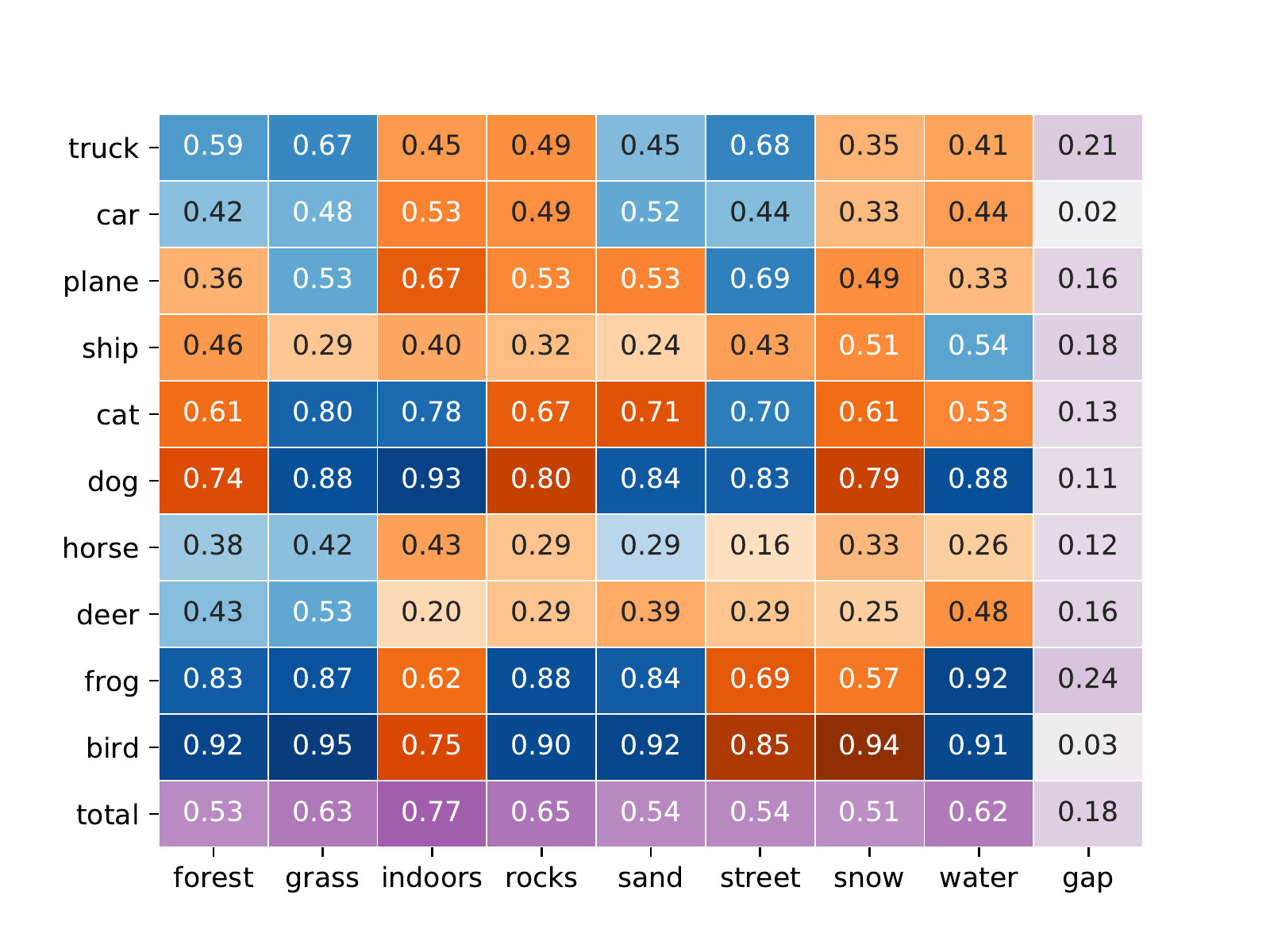}
\caption{Category vs Location}
\end{subfigure}

\caption{Accuracy of ResNext-50 (32x4d) for all combinations of classes and attributes.}

\end{figure*}

\end{document}